\documentclass[11pt]{article}

\usepackage[final]{acl}

\usepackage{times}
\usepackage{latexsym}
\usepackage{multirow}
\usepackage{amsmath}
\usepackage{textcomp}
\usepackage{listings}
\usepackage{fontawesome5}

\newcounter{promptcounter}
\usepackage[most]{tcolorbox}
\usepackage[normalem]{ulem}

\usepackage[T1]{fontenc}

\usepackage[utf8]{inputenc}

\usepackage{microtype}

\usepackage{inconsolata}

\usepackage{graphicx}

\usepackage{placeins}
\usepackage{contour}
\usepackage{booktabs}
\usepackage{enumitem}
\newcommand{\AraSEG}{\textbf{\textsc{AraSEG}}}
\newcommand{\hide}[1]{}

\newcommand{\THA}{{$\theta$}}

\newcommand{\SHADDA}{{$\sim$}}

\usepackage{arabtex}

\usepackage{utf8}
\usepackage{epstopdf}

\title{Arabic Sentence Segmentation \\ Across Genres and Punctuation Conditions}

\author{Mohammed Elkholy\textsuperscript{1} \quad Khalid N. Elmadani\textsuperscript{1,2} \quad \textbf{Nizar Habash\textsuperscript{1,2}} \quad Bashar Alhafni\textsuperscript{1}\\ \textsuperscript{1}Mohamed bin Zayed University of Artificial Intelligence \\ \textsuperscript{2}New York University Abu Dhabi\\ \texttt{\{mohammed.elkholy,khalid.elmadani,bashar.alhafni\}@mbzuai.ac.ae}\\ \texttt{nizar.habash@nyu.edu}}

\begin{document}
\setcode{utf8}
\vocalize

\maketitle
\begin{abstract}
Sentence segmentation in Arabic is challenging due to ambiguous and inconsistent punctuation, with many texts lacking reliable sentence boundary markers. Existing approaches rely heavily on punctuation cues and are typically evaluated on well-formed text, limiting their robustness in realistic Arabic settings. To address this, we introduce {\AraSEG}, a genre-diverse sentence segmentation corpus spanning eight genres and a wide range of punctuation and document structure conditions. Using {\AraSEG}, we evaluate LLMs, lightweight encoder models, and dependency parser-based models under increasingly challenging segmentation settings. Our experiments show that lightweight encoders, and even dependency parser-based models, outperform LLMs under the hardest conditions. We further investigate the effects of training data size and genre diversity, finding that performance eventually saturates and cross-genre generalization remains challenging. We also demonstrate that accurate sentence segmentation substantially improves downstream dependency parsing. We make our code, data, and models publicly available.\footnote{\url{https://github.com/mbzuai-nlp/araseg}}

\end{list} 
\end{abstract}

\section{Introduction}
Sentences are fundamental units of processing in NLP. Most datasets are annotated at the sentence level, and many downstream systems, including dependency parsers, NER systems, and retrieval pipelines, assume sentence-delimited input. Sentence segmentation is therefore a critical preprocessing step for raw text. It is also increasingly important for LLM pipelines, where sentence boundaries improve text chunking for pretraining, alignment~\cite{qiu2025sentencelevelrewardmodelgeneralize}, and training efficiency~\cite{zheng2025groupsequencepolicyoptimization}, even in models with long context windows~\cite{liu-etal-2026-think}. 

\begin{figure}[t!]
    \centering
    \includegraphics[width=\linewidth]{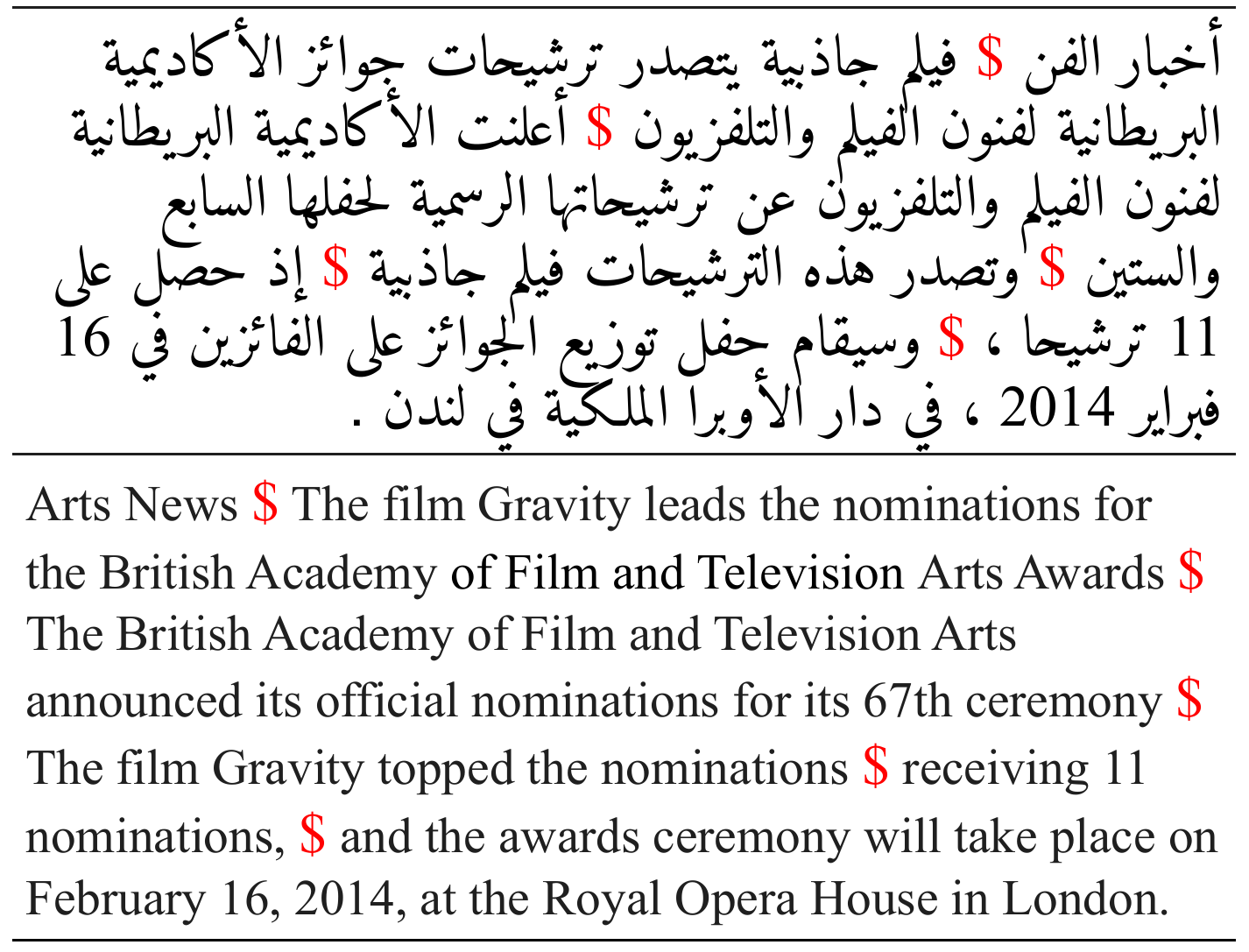}
    \caption{Example Arabic paragraph annotated with sentence segmentation boundaries \textcolor{red}{\$}.}
    \label{fig:intro-example}
\end{figure}

Despite its importance, sentence segmentation remains relatively underexplored for Arabic, with limited benchmarks and evaluation settings for systematically studying the task. Unlike many languages, Arabic often exhibits sparse, inconsistent, or entirely absent punctuation, particularly in historical and literary texts predating the widespread adoption of modern punctuation. Instead, clauses and sentences are often linked through coordinating conjunctions and discourse markers, making sentence boundaries less explicit and reducing the reliability of punctuation-based segmentation.
%
Figure~\ref{fig:intro-example} illustrates an example of Arabic sentence segmentation, where sentence boundaries are not always recoverable from punctuation alone.

To address this, we introduce {\AraSEG}, a manually annotated, genre-diverse benchmark for Arabic sentence segmentation spanning eight genres with varying writing styles, punctuation usage, and document structures. Using {\AraSEG}, we benchmark LLMs, lightweight encoder models, and dependency parser-based approaches across segmentation settings with varying punctuation and document conditions. Our experiments show that lightweight supervised models substantially outperform LLMs. Our contributions are as follows:

\begin{enumerate}
    \item We introduce {\AraSEG}, the first genre-diverse benchmark for Arabic sentence segmentation.
    \item We benchmark lightweight encoder, dependency parser-based, and LLM approaches across multiple punctuation and document settings, showing that lightweight supervised models substantially outperform LLMs.
    \item We analyze punctuation ambiguity, training data size, and cross-genre generalization, and the impact of sentence segmentation on downstream dependency parsing.
\end{enumerate}

\section{Background and Related Work}
\subsection{Sentence Segmentation}
Several approaches to sentence segmentation have been explored in prior work. Early methods relied on punctuation and handcrafted heuristics, as seen in rule-based systems such as SpaCy\textsubscript{SENT}~\cite{honnibal2020spacy} and PySBD~\cite{sadvilkar-neumann-2020-pysbd}, while unsupervised methods such as Punkt~\cite{kissandstrunk2006} used statistical patterns in raw text to identify sentence boundaries. More recently, self-supervised approaches including Ersatz, Where's the Point (WtP), and Segment Any Text (SaT)~\cite{wicks-post-2021-unified,minixhofer-etal-2023-wheres,frohmann-etal-2024-segment} have leveraged transformer encoders for multilingual sentence segmentation. Recent work has also explored segmentation with LLMs~\cite{retkowski-etal-2026-paragraph}, though encoder-based models continue to outperform LLMs in historically grounded and low-resource settings~\cite{bilitski-etal-2026-automatic}. While many of these approaches support multilingual segmentation and include Arabic in their evaluations, they do not examine the linguistic and structural challenges specific to Arabic sentence segmentation.



Prior work on Arabic text segmentation has explored clause segmentation~\cite{touir2008semantic,keskes-etal-2012-clause}, semantic chunking~\cite{computation13060151,chirkunov-etal-2026-linear}, and punctuation restoration~\cite{app122010559}. However, little work has addressed sentence segmentation directly; closest to ours, \citet{Mekki2022} studied segmentation for Tunisian Arabic. The task therefore remains underexplored across genres and punctuation/document conditions. To the best of our knowledge, our work presents the first genre-diverse, manually annotated Arabic sentence segmentation corpus for systematic evaluation across these settings.
\subsection{Arabic Linguistic Challenges} 
\label{sec:arabic-facts}
Arabic sentence segmentation is particularly challenging due to the ambiguous and inconsistent use of punctuation. Modern punctuation was introduced into Arabic relatively late alongside the standardization of Modern Standard Arabic (MSA), and many historical or literary texts therefore contain little to no punctuation. Instead of relying on explicit sentence markers, Arabic frequently uses coordinating conjunctions and discourse markers such as \<و> \textit{wa}\footnote{Arabic HSB transliteration~\cite{Habash:2007:arabic-transliteration}.} `and', \<ف> \textit{fa} `so/then', and \<ثم> \textit{{\THA}um{\SHADDA}a} `then/afterwards'  to connect clauses and sentences. Even when punctuation is present, its function is highly ambiguous and varies across genres and writing styles. For example, the Arabic comma (\<،>) may act either as an intra-sentential connector or as a sentence boundary marker~\cite{hadrich-belguith-etal-2005-segmentation,zaghouani2016toward,yagi2024}. Prior work has also shown that punctuation misuse is a major source of grammatical errors in Arabic writing~\cite{zaghouani-etal-2014-large}. These characteristics make punctuation a substantially less reliable signal for sentence segmentation in Arabic than in many other languages.

\section{Dataset}
\subsection{Overview}
{\AraSEG} is a large-scale Arabic sentence segmentation corpus covering a diverse range of genres and writing styles. It consists of 2,187 documents containing 48,250 paragraphs and 121,623 sentences, totaling 1.45M tokens from 30 sources spanning 8 genres. This diversity allows {\AraSEG} to capture substantial variation in Arabic sentence boundary realization, including punctuated and unpunctuated text as well as inconsistent and ambiguous punctuation usage. The corpus provides word-level sentence boundary labels, enabling the study of robust Arabic sentence segmentation across diverse domains and writing conditions.

\begin{table*}[ht]
\centering
\setlength{\tabcolsep}{3.5pt}
\begin{tabular}{lccccccccc}
\toprule
& \textbf{\#Docs} & \textbf{\#Paras} & \textbf{\#Sents} & \textbf{\#Tokens} & \textbf{\#Words} & \textbf{\#Pnx} & \textbf{\#PnxClus} & \textbf{PnxDen} & \textbf{Sent. Len.} \\
\midrule
Children   & 422  & 9.8K  & 16.7K  & 172.7K & 144.8K & 35.6K  & 27.8K  & 20.6\% & 10.4 \\
Education  & 605  & 18.8K & 50.3K  & 373.3K & 288.5K & 94.6K  & 84.7K  & 25.3\% & 7.4 \\
Literature & 406  & 8.0K  & 21.0K  & 339.4K & 298.0K & 48.1K  & 41.5K  & 14.2\% & 16.2 \\
Media      & 101  & 1.2K  & 2.5K   & 32.9K  & 29.3K  & 4.1K   & 3.6K   & 12.4\% & 13.2 \\
Poetry     & 30   & 1.7K  & 1.7K   & 11.7K  & 10.9K  & 0.9K   & 0.8K   & 7.7\%  & 6.9 \\
Politics   & 17   & 5.9K  & 9.7K   & 147.0K & 131.7K & 15.5K  & 15.2K  & 10.6\% & 15.1 \\
Religion   & 438  & 1.4K  & 13.7K  & 216.0K & 185.1K & 34.8K  & 30.9K  & 16.1\% & 15.7 \\
Wiki       & 168  & 1.6K  & 6.1K   & 128.6K & 113.1K & 17.7K  & 15.5K  & 13.8\% & 21.2 \\
\midrule\midrule
Total & 2.2K & 48.3K & 121.6K & 1.4M & 1.2M & 251.3K & 220.1K & 17.7\% & 11.7 \\
\bottomrule
\end{tabular}
\caption{Genre-level statistics for {\AraSEG}, including documents (Docs), paragraphs (Paras), sentences (Sents), total tokens (Tokens), word tokens (Words), punctuation tokens (Pnx), punctuation clusters (PnxClus), punctuation density (PnxDen), and average sentence length (Sent. Len.).}
\label{tab:genre_stats_summary}
\end{table*}


\subsection{Corpus Construction}
{\AraSEG} is constructed from multiple existing Arabic corpora spanning a diverse set of domains and genres. The primary sources of the dataset are the CAMeL Treebank (CAMeLTB) \cite{habash-etal-2022-camel} and the Balanced Arabic Readability Evaluation Corpus (BAREC) \cite{elmadani-etal-2025-large}. Although originally developed for dependency parsing and Arabic readability assessment, respectively, these resources include manually curated sentence segmentations and paragraph boundaries, making them well-suited for sentence segmentation research. We provide brief descriptions of all genres and sources in Appendix~\ref{sec:appendix_dataset_sources}.

To further increase genre diversity and capture variation in sentence segmentation, particularly in religious text, we additionally include the entire Quran, along with approximately 45K tokens from the Old Testament (OT) and 45K tokens from the New Testament (NT) beyond the portions already present in CAMeLTB.

\paragraph{Corpus Splits} The corpus is split at the document level to prevent data leakage. We retain the original CAMeLTB and BAREC splits and assign new splits to the added Quran, OT, and NT documents, greedily optimizing for word count. Overall, the dataset is split by word count into approximately \textbf{Train ($\simeq$80\%)}, \textbf{Dev ($\simeq$10\%)}, and \textbf{Test ($\simeq$10\%)}. Appendix~\ref{sec:dataset_splits} shows the number of documents, paragraphs, sentences and tokens per split.


\subsection{Annotations}
We re-purpose the gold sentence segmentations provided in CAMeLTB and BAREC for our task. Both datasets follow the annotation guidelines introduced by CAMeLTB, which begins with automatic segmentation based on punctuation marks (,.;?!:), followed by manual correction through controlled splitting and merging operations. Splits are introduced only when adjacent spans form independent sentences, while merges are applied when punctuation incorrectly separates parts of a single sentence. Dependent or incomplete clauses are therefore never segmented into separate sentences.

For the Quran, OT, NT, and poetry, verse boundaries serve as sentence boundaries following CAMeLTB and Arabic Poetry Treebank (ARPoT) \cite{al-ghamdi-etal-2021-dependency}; in poetry, they also serve as paragraph boundaries. While verses and grammatical sentences in the Quran do not map one-to-one, we intentionally retained the verse segmentation for consistency with existing Arabic treebanks and Quranic resources. For sources without explicit paragraph annotations, we derive paragraph boundaries from the document structure, including Quran chapters and ArabicMMLU questions. In ArabicMMLU, the question prompt and each answer option are treated as separate sentences.

\subsection{Corpus Statistics}
Table~\ref{tab:genre_stats_summary} summarizes corpus statistics across genres, including the number of documents, paragraphs, sentences, total tokens, word tokens, punctuation (Pnx) tokens, and punctuation clusters. We define a punctuation cluster as a contiguous sequence of one or more punctuation marks, with individual punctuation marks treated as singleton clusters. We additionally report \textit{Pnx Density}, computed as the proportion of punctuation tokens relative to the total number of tokens in each genre. Detailed source-level dataset statistics are provided in Appendix~\ref{sec:appendix_dataset_stats}.

{\AraSEG} exhibits substantial variation across genres in corpus size, sentence length, and punctuation usage. Expository genres such as Wiki, literature, and religious text tend to contain substantially longer sentences, while educational text and poetry often exhibit shorter or more fragmented structures. Punctuation usage is similarly inconsistent across genres, reflecting the noisy nature of punctuation in diverse Arabic text (\S\ref{sec:arabic-facts}).

\subsection{Sentence Boundary Realization}
\label{sec:pnx-reliability}
We analyze sentence boundary realization in {\AraSEG} through the distribution of punctuation- and word-based segmentation, and the reliability of punctuation as a sentence boundary indicator.

\begin{table}[t!]
\centering
\setlength{\tabcolsep}{4pt}
\begin{tabular}{lc@{\hspace{3pt}}cc@{\hspace{3pt}}cc}
\toprule
& \multicolumn{2}{c}{\textbf{Pnx Segs}} & \multicolumn{2}{c}{\textbf{Word Segs}} & \textbf{Total} \\
\midrule
Children   & 13k  & (76\%) & 4k   & (24\%) & 17k \\
Education  & 21k  & (41\%)  & 29k  & (59\%) & 50k \\
Literature & 18k  & (84\%) & 3k   & (16\%) & 21k \\
Media      & 2k   & (80\%) & 0.5k  & (20\%) & 3k \\
Poetry     & 0.2k  & (12\%) & 2k   & (88\%) & 2k \\
Politics   & 7k   & (68\%) & 3k   & (32\%) & 10k \\
Religion   & 8k   & (60\%) & 6k   & (40\%) & 14k \\
Wiki       & 6k   & (97\%) & 0.2k  & (3\%)  & 6k \\
\midrule\midrule
Total & 74k & (61\%) & 48k & (39\%) & 122k \\
\bottomrule
\end{tabular}
\caption{Distribution of punctuation (Pnx)- and word-based segmentation across genres in \textbf{\AraSEG}.}
\label{tab:word-seg}
\end{table}

					
\paragraph{Word vs. Punctuation Segmentation}
\label{para:word_vs_pnx}
We distinguish \textit{punctuation-based} segmentation, where sentence boundaries are marked using explicit punctuation, from \textit{word-based} segmentation, where boundaries occur without sentence-final punctuation. Table~\ref{tab:word-seg} shows that word-based segmentation accounts for 39\% of sentence boundaries in {\AraSEG}. This distribution also varies substantially across genres: poetry and educational text rely heavily on word-based segmentation, while Wiki and literature predominantly use punctuation-based boundaries.


\paragraph{Punctuation Segmentation Precision} We define segmentation precision for punctuation cluster $p$ as:
\[
\text{SegPrecision}(p) =
\frac{N_{\text{seg}}(p)}{N_{\text{occ}}(p)}
\]
where $N_{\text{seg}}(p)$ denotes the number of occurrences of $p$ that mark a sentence boundary, and $N_{\text{occ}}(p)$ denotes its total number of occurrences. Table~\ref{tab:pnx-genre-precision} shows that segmentation precision varies substantially across both punctuation types and genres. Canonical sentence-final punctuation such as ?, !, and . generally exhibits high precision overall, whereas punctuation marks such as commas (\<،>), quotation marks ("), and hyphens (-) are highly ambiguous boundary indicators. Precision also varies considerably across genres: periods achieve high precision in Wiki and literary text but are substantially less reliable in poetry and religious text, reflecting genre-specific punctuation conventions and usage patterns. Overall counts for most frequent punctuations are shown in Appendix~\ref{sec:app_overall_pnx}.

\begin{table*}[!t]
\centering
\setlength{\tabcolsep}{6pt}
\begin{tabular}{lccccccccccc}
\toprule
 & \textbf{,} & \textbf{.} & \textbf{(} & \textbf{)} & \textbf{:} & \textbf{"} & \textbf{?} & \textbf{-} & \textbf{..} & \textbf{;} & \textbf{!} \\
\midrule
Children & 22.5 & 97.6 & 0 & 18.2 & 11.2 & 3.9 & 99.1 & 1.7 & 70.0 & 57.0 & 98.0 \\
Education & 23.5 & 92.5 & 0 & 1.5 & 72.1 & 9.4 & 95.7 & 1.4 & 76.1 & 21.9 & 94.1 \\
Literature & 21.9 & 98.7 & 0 & 7.7 & 11.9 & 2.2 & 98.5 & 2.0 & 84.6 & 47.5 & 98.0 \\
Media & 37.7 & 98.4 & 0 & 16.9 & 10.0 & 4.6 & 100.0 & 4.8 & 98.3 & 40.0 & 95.8 \\
Poetry & 0.0 & 39.7 & 0 & 0 & 26.7 & 17.0 & 84.3 & 0 & 2.6 & -- & 100.0 \\
Politics & 5.3 & 86.0 & 0 & 13.8 & 80.4 & 2.3 & -- & 0 & -- & 33.9 & -- \\
Religion & 8.6 & 15.5 & 0 & 100.0 & 0.4 & 2.2 & 0.2 & 0 & -- & 0 & 2.3 \\
Wiki & 18.0 & 94.2 & 0 & 3.5 & 9.8 & 1.0 & 94.4 & 0.6 & 100.0 & 47.6 & 50.0 \\
\midrule\midrule
Overall & 18.6 & 82.9 & 0 & 20.5 & 45.2 & 3.4 & 89.3 & 1.6 & 70.0 & 41.9 & 83.7 \\
\bottomrule
\end{tabular}
\caption{Segmentation precision of frequent punctuation across {\AraSEG} genres. Punctuation with fewer than 50 occurrences are omitted.}
\label{tab:pnx-genre-precision}
\end{table*}

\section{Experimental Setup}
We formulate Arabic sentence segmentation as a binary token classification task, where the goal is to predict whether a sentence boundary follows each token. Below, we describe the task variants, models, and evaluation metrics.


\subsection{Task Variants}
To investigate the role of punctuation and document structure in Arabic sentence segmentation, we train and evaluate models under four task variants that differ by the availability of punctuation and paragraph boundary information. Paragraph boundaries correspond to the natural structure of each source in {\AraSEG}. The four  variants are:
\begin{itemize}
    \item \textbf{No-Punctuation No-Paragraph (NoPnx-NP)}: punctuation, domain-specific boundary markers (e.g., verse numbers in Quran/OT/NT and question/multiple-choice markers in ArabicMMLU), and paragraph boundaries are removed.
    \item \textbf{No-Punctuation Paragraph-Aware (NoPnx-PA)}: punctuation and domain-specific boundary markers are removed, while paragraph boundaries are retained.
    \item \textbf{No-Paragraph (NP)}: punctuation is retained, but paragraph boundaries are removed.
    \item \textbf{Paragraph-Aware (PA)}: punctuation and paragraph boundaries are retained.
\end{itemize}




\subsection{Models}
We evaluate three approaches to Arabic sentence segmentation: lightweight encoder models, dependency parser-based models, and LLMs. Training details are provided in Appendix~\ref{sec:appendix_training}.

\paragraph{Baselines} 
We compare against rule-based, simple frequency-based and neural sentence segmentation baselines. For rule-based approaches, we evaluate SpaCy\textsubscript{SENT} and PySBD, alongside an unsupervised NLTK Punkt segmenter trained on {\AraSEG}. We also include three simple, interpretable frequency-based baselines trained on {\AraSEG}: (1) Pnx-Boundary, which predicts boundaries using empirical punctuation end-of-sentence probabilities; (2) Lexical-Boundary, which predicts the probability that a word begins a sentence, motivated by the frequent use of conjunctions at sentence onset in Arabic; and (3) ENS$_\text{Pnx+Lexical}$, which predicts a boundary when either baseline does so. Finally, we compare against recent transformer-based models, namely Ersatz and the largest model from the SaT family.

\paragraph{BERT Models} 
We fine-tune CAMeLBERT-MSA~\cite{inoue-etal-2021-interplay} for token classification, using the final subtoken representation of each word for boundary prediction. Due to the model’s maximum context length of 512 tokens, long documents are processed using overlapping sliding-window chunks, with logits averaged for tokens appearing in multiple windows.

To better model the variability of Arabic punctuation usage, we explore several task-specific training strategies. For non-punctuated settings, we introduce auxiliary punctuation insertion ($\text{CAMeLBERT}_{\text{PnxInsert}}$), while for punctuated settings we apply punctuation dropout ($\text{CAMeLBERT}_{\text{PnxDrop}}$) to improve robustness to noisy or inconsistent punctuation. 
Additionally, for punctuated settings (NP and PA), we train specialized models that predict either word-based boundaries ($\text{CAMeLBERT}_{\text{Word}}$) or punctuation-based boundaries ($\text{CAMeLBERT}_{\text{Pnx}}$), and ensemble their predictions (ENS-$\text{CAMeLBERT}_{\text{Pnx+Word}}$).



\paragraph{Dependency Parsers}
We investigate whether dependency parsing can be leveraged for sentence segmentation by training biaffine dependency parsers~\cite{DBLP:journals/corr/DozatM16} on silver-standard annotations generated using CamelParser 2.0~\cite{elshabrawy-etal-2023-camelparser2}. Since dependency parsers are typically applied to pre-segmented sentences and do not explicitly model sentence boundaries, we adapt dependency parsing to operate over entire documents. Specifically, we introduce an artificial document-root token at the beginning of each document and attach the root of every sentence to this node, producing a single document-level dependency tree. At inference time, sentence boundaries are recovered by traversing the subtrees rooted at the document root and assigning a boundary after the rightmost token in each subtree.

In addition to the original dependency trees (Full Orig Deprel), we construct flattened trees in which each token attaches to the immediately preceding token. This is motivated by the observation that roughly 50\% of tokens in the original trees already attach to the preceding token, meaning that many local attachments remain unchanged after flattening. For the flattened trees, we consider three dependency-relations settings: retaining the original dependency relations (Flat Orig Deprel), removing dependency relations (Flat No Deprel), and retaining only direct dependency relations (Flat Direct Deprel). These variants allow us to assess how much sentence segmentation benefits from dependency labels and full tree structure.



\begin{table*}[ht]
\centering
\setlength{\tabcolsep}{2.3pt}
\begin{tabular}{lccc|ccc|ccc|ccc|c}

\toprule
& \multicolumn{3}{c}{\textbf{NoPnx-NP}} 
& \multicolumn{3}{c}{\textbf{NoPnx-PA}}
& \multicolumn{3}{c}{\textbf{NP}}
& \multicolumn{3}{c}{\textbf{PA}}
& \textbf{Avg} \\
\cmidrule(lr){2-4}
\cmidrule(lr){5-7}
\cmidrule(lr){8-10}
\cmidrule(lr){11-13}
\cmidrule(lr){14-14}
& \textbf{P} & \textbf{R} & \textbf{F\textsubscript{1}}
& \textbf{P} & \textbf{R} & \textbf{F\textsubscript{1}}
& \textbf{P} & \textbf{R} & \textbf{F\textsubscript{1}}
& \textbf{P} & \textbf{R} & \textbf{F\textsubscript{1}}
& \textbf{F\textsubscript{1}} \\
\midrule

Paragraph Breaks
& -- & -- & --
& \textbf{100.0} & 33.7 & 45.2
& -- & -- & --
& \textbf{100.0} & 33.6 & 45.2
& 45.2 \\
\midrule

Pnx-Boundary & -- & -- & -- & -- & -- & -- & 70.8 & 58.3 & 60.6 & 75.5 & 71.5 & 70.9 & --  \\
Lexical-Boundary & 48.5 & 29.4 & 32.9 & 72.7 & 48.9 & 55.5 & 45.7 & 35.5 & 34.7 & 70.5 & 53.7 & 57.2 & 45.0 \\
ENS$_\text{Pnx+Lexical}$ & 48.5 & 29.4 & 32.9 & 72.7 & 48.9 & 55.5 & 50.6 & 67.6 & 55.4 & 61.9 & 78.1 & 67.3 & 52.8 \\

NLTK Punkt
& \textbf{100.0} & 3.7 & 6.8
& \textbf{100.0} & 33.7 & 45.2
& 81.2 & 47.6 & 52.3
& 85.8 & 62.7 & 65.2
& 42.4 \\

SpaCy\textsubscript{SENT}
& \textbf{100.0} & 3.7 & 6.8
& \textbf{100.0} & 33.7 & 45.2
& 82.1 & 48.8 & 54.3
& 89.8 & 62.8 & 67.3
& 43.4 \\

Ersatz
& \textbf{100.0} & 3.7 & 6.8
& \textbf{100.0} & 33.7 & 45.2
& 80.2 & 46.9 & 51.6
& 85.0 & 62.4 & 64.7
& 42.1 \\

PySBD
& \textbf{100.0} & 3.7 & 6.8
& \textbf{100.0} & 33.7 & 45.2
& 77.2 & 48.7 & 52.1
& 80.7 & 62.1 & 63.1
& 41.8 \\

SaT
& 88.2 & 43.2 & 52.5
& 89.7 & 55.6 & 64.9
& 90.2 & 55.7 & 60.9
& 93.8 & 58.4 & 66.9
& 61.3 \\
\midrule

Fanar-2-27B-Instruct
&  95.6 & 24.0 & 32.3
& 90.8 & 47.9 & 57.2
& 80.0 & 24.4 & 30.9
& 90.1 & 70.8 & 74.7
& 48.8 \\

Qwen3.6-27B
& 92.3 & 57.4 & 61.2
& 84.6 & 73.7 & 75.1
& 92.7 & 71.1 & 74.7
& 85.5 & 83.6 & 81.7
& 73.2 \\

Jais-2-70B-Chat
& 85.0 & 50.7 & 48.1
& 65.6 & 77.4 & 65.8
& 92.0 & 53.9 & 57.7
& 69.6 & 75.2 & 67.5
& 59.8 \\

Qwen2.5-72B-Instruct
& 89.6 & 63.7 & 68.4
& 83.1 & 82.2 & 80.5
& 94.6 & 62.3 & 69.3
& 85.7 & 82.5 & 81.2
& 74.8 \\

GPT-4.1
& {98.2} & 25.4 & 30.6
& {97.3} & 62.4 & 70.7
& \textbf{96.3} & 57.9 & 66.0
& 96.6 & 69.4 & 76.0
& 60.8 \\

GPT-5.5
& 96.4 & 41.5 & 47.6
& 95.7 & 70.1 & 76.4
& 90.9 & 60.2 & 66.8
& 94.7 & 70.5 & 76.2
& 66.8 \\

Gemini-3.1-Pro
& 95.3 & 67.1 & 74.8
& 95.4 & 72.6 & 79.2
& 91.6 & 62.6 & 69.5
& 94.4 & 71.0 & 76.8
& 75.1 \\
\midrule

Full Orig Deprel
& 72.3 & 70.8 & 70.6
& 79.5 & 79.0 & 78.5
& 84.7 & 88.2 & 86.0
& 89.9 & 93.0 & 91.2
& 81.6 \\

Flat No Deprel
& 84.3 & 65.6 & 72.9
& 87.3 & 78.3 & 82.0
& 93.6 & 88.2 & 90.3
& 95.3 & 92.9 & 93.8
& 84.7 \\

Flat Direct Deprel
& 84.2 & 67.9 & 74.2
& 88.3 & 79.5 & 83.0
& 92.9 & 89.4 & 90.6
& 95.5 & 94.0 & 94.5
& 85.6 \\

Flat Orig Deprel
& 84.0 & 69.8 & 75.4
& 88.4 & {79.6} & {83.2}
& 93.4 & 89.5 & 91.0
& 95.4 & 94.5 & 94.7
& {86.1} \\
\midrule
CAMeLBERT
& 85.5 & \textbf{84.9} & \textbf{84.8}
& 89.9 & \textbf{87.4} & \textbf{88.3}
& {94.1} & {93.2} & \textbf{93.4}
& 96.3 & \textbf{95.3} & \textbf{95.6}
& \textbf{90.5} \\

$\text{CAMeLBERT}_{\text{PnxDrop}}$
& -- & -- & --
& -- & -- & --
& 93.5 & 93.1 & 93.1
& 94.9 & 94.7 & 94.6
& -- \\

$\text{CAMeLBERT}_{\text{PnxInsert}}$
& 85.4 & {84.7} & {84.6}
& 88.8 & 79.2 & 82.1
& -- & -- & --
& -- & -- & --
& -- \\

$\text{CAMeLBERT}_{\text{Pnx}}$
& -- & -- & --
& -- & -- & --
& 84.8 & 62.0 & 67.6
& 86.0 & 62.6 & 68.4
& -- \\

$\text{CAMeLBERT}_{\text{Word}}$
& -- & -- & --
& -- & -- & --
& 79.0 & 31.7 & 38.2
& 84.5 & 32.4 & 39.3
& -- \\

ENS-$\text{CAMeLBERT}_{\text{Pnx+Word}}$
& -- & -- & --
& -- & -- & --
& 93.4 & \textbf{93.7} & 93.3
& 96.4 & 95.0 & 95.5
& -- \\
\bottomrule
\end{tabular}
\caption{Results on the Dev set of {\AraSEG} across the four task variants. Best results are in \textbf{bold}.}
\label{tab:dev-results}
\end{table*}
\paragraph{LLMs} We benchmark three commercial LLMs: GPT-5.5~\cite{gpt5}, Gemini-3.1-Pro~\cite{gemini}, and GPT-4.1~\cite{openai2024gpt4technicalreport}, as well as four open-source LLMs: two Arabic-centric models, Fanar-2-27B-Instruct~\cite{fanarteam2026fanar20arabicgenerative} and Jais-2-70B-Chat~\cite{anwar2026jais2familyarabiccentric}, and two multilingual models, Qwen3.6-27B \cite{qwen3.6-27b} and Qwen2.5-72B-Instruct \cite{qwen2}.
Rather than formulating sentence segmentation as binary token classification, we prompt LLMs to regenerate the input with sentence boundaries explicitly marked. We use this formulation because LLMs are prone to hallucinations and often struggle with counting-based tasks~\cite{xu-ma-2025-llm}.

Since outputs may contain omitted, inserted, or modified tokens and punctuation, we align them with the input using the algorithm proposed by \newcite{alhafni-etal-2023-advancements} before evaluation. Documents exceeding model context limits are processed using overlapping chunks to maintain segmentation consistency. Prompts, chunking details, and licenses are in Appendices~\ref{sec:llm_prompts}, \ref{sec:appendix_llms_chunking}, and~\ref{app:license}, respectively.


\begin{table*}[t]
\centering
\setlength{\tabcolsep}{4pt}
\begin{tabular}{l ccc | ccc | ccc | ccc | c}
\toprule
& \multicolumn{3}{c}{\textbf{NoPnx-NP}} 
& \multicolumn{3}{c}{\textbf{NoPnx-PA}}
& \multicolumn{3}{c}{\textbf{NP}}
& \multicolumn{3}{c}{\textbf{PA}}
& \textbf{Avg} \\
\cmidrule(lr){2-4}
\cmidrule(lr){5-7}
\cmidrule(lr){8-10}
\cmidrule(lr){11-13}
\cmidrule(lr){14-14}

& \textbf{P} & \textbf{R} & \textbf{F\textsubscript{1}}
& \textbf{P} & \textbf{R} & \textbf{F\textsubscript{1}}
& \textbf{P} & \textbf{R} & \textbf{F\textsubscript{1}}
& \textbf{P} & \textbf{R} & \textbf{F\textsubscript{1}}
& \textbf{F\textsubscript{1}} \\
\midrule

Paragraph Breaks
& -- & -- & --
& \textbf{100.0} & 35.1 & 46.4
& -- & -- & --
& \textbf{100.0} & 35.1 & 46.5
& 46.5 \\
\midrule





SaT
& 90.4 & 44.0 & 53.0
& 92.0 & 54.6 & 63.7
& 92.8 & 54.3 & 61.8
& 94.8 & 57.8 & 66.8
& 61.3 \\
\midrule




Gemini-3.1-Pro
& \textbf{95.2} & 67.8 & 75.3
& 96.6 & 73.2 & 79.6
& {94.4} & 60.6 & 69.5
& 96.6 & 69.2 & 76.6
& 75.2 \\
\midrule
Flat Orig Deprel
& 85.6 & {70.9} & {76.5}
& 88.8 & {79.5} & {82.8}
& 93.5 & {89.0} & {90.7}
& 88.8 & {79.5} & {82.8}
& {83.2} \\\midrule

CAMeLBERT
& 86.8 & \textbf{86.2} & \textbf{85.9}
& 90.4 & \textbf{88.0} & \textbf{88.4}
& 94.0 & \textbf{93.0} & \textbf{93.2}
& 96.1 & \textbf{94.7} & \textbf{95.2}
& \textbf{90.7} \\
\bottomrule
\end{tabular}
\caption{Results on the Test set of {\AraSEG} across the four task variants. Best results are in \textbf{bold}.}
\label{tab:test-results}
\end{table*}

\subsection{Evaluation Metrics}
We evaluate models using boundary-level precision (P), recall (R), and F\textsubscript{1}, where predictions are correct only if they exactly match gold boundaries. We report macro-averaged document-level F\textsubscript{1} to avoid bias toward larger documents. We do not report P\textsubscript{k}~\cite{Pk} or WinDiff~\cite{pevzner-hearst-2002-critique}, which target coarse-grained segmentation rather than sentence boundary detection.

\section{Results}
Table~\ref{tab:dev-results} presents results on the Dev set.


\paragraph{Baselines} The paragraph-break baseline achieves identical performance in the PA settings, recovering only paragraph boundaries. 
Similarly, punctuation-based baselines: NLTK Punkt, SpaCy\textsubscript{SENT}, Ersatz, and PySBD, achieve identical performance in NoPnx-NP and NoPnx-PA. Without punctuation, these systems recover only document-final boundaries, yielding perfect precision but very low recall. Their performance improves substantially when punctuation is available, reaching up to 54.3 F\textsubscript{1} in NP and 67.3 F\textsubscript{1} in PA, with SaT being the strongest baseline on average (61.3 F\textsubscript{1}).

The corpus-derived Pnx-Boundary baseline outperforms fixed punctuation baselines and is competitive with SaT in NP (-0.3 F\textsubscript{1}) and surpasses it in PA (+4.0 F\textsubscript{1}), highlighting the limitations of fixed punctuation rules across diverse sources and genres. Although Lexical-Boundary identifies some sentence boundaries in the NoPnx settings, its recall remains limited (29\% in NoPnx-NP; 49\% in NoPnx-PA). Combining lexical and punctuation cues (ENS$_\text{Pnx+Lexical}$) improves over Lexical-Boundary in NP and PA (55.4 and 67.3 F\textsubscript{1}, respectively) but remains below Pnx-Boundary, suggesting that lexical cues introduce additional false-positive boundaries when punctuation is available.

\paragraph{LLMs}
Gemini-3.1-Pro is the strongest LLM across all settings. All LLMs benefit from paragraph boundaries and punctuation, but the gains vary substantially across models. GPT-4.1 benefits the most from paragraph boundaries in the NoPnx settings (+40 F\textsubscript{1}) and from punctuation in NP (+35 F\textsubscript{1}), while Fanar exhibits the largest gains in PA from both paragraph boundaries (+44 F\textsubscript{1}) and punctuation (+18 F\textsubscript{1}). In contrast, Gemini is the only model that does not consistently benefit from punctuation and even exhibits performance decreases when punctuation is present. Notably, the multilingual Qwen models substantially outperform Arabic-centric models of comparable size. Qwen3.6-27B outperforms Fanar-2-27B by 24.4 F\textsubscript{1} points on average, while Qwen2.5-72B-Instruct surpasses Jais-2-70B by 15.0 F\textsubscript{1} points. Across all settings, LLMs generally exhibit high precision but lower recall, indicating a tendency to under-segment rather than over-segment text.


\paragraph{Dependency Parsers}
All parser-based variants outperform LLMs. Flattening the dependency trees consistently improves performance over the full parser, with Flat Orig Deprel achieving the best average F\textsubscript{1}. All parser variants benefit from both paragraph boundaries and punctuation. The effect of paragraph boundaries is more pronounced in the NoPnx settings (+8 F\textsubscript{1} on average), while punctuation provides larger gains in the NP settings (+15 F\textsubscript{1} on average).


\paragraph{CAMeLBERT} CAMeLBERT achieves the best overall performance (average 90.5 F\textsubscript{1}), outperforming all parsers and LLMs. Auxiliary punctuation insertion in the NoPnx settings and punctuation dropout in the punctuated settings provide no gains over the base model. In punctuated settings, punctuation-only and word-only variants perform substantially worse than the jointly trained model, suggesting that both signals are needed for accurate segmentation. However, ensembling the two specialized models nearly recovers the performance of the jointly trained model within 0.1 F\textsubscript{1}.

\paragraph{Test Results} Table~\ref{tab:test-results} reports the best system from each model family on the Test set. Consistent with the Dev results, CAMeLBERT remains the strongest system, achieving an average F\textsubscript{1} of 90.7. Table~\ref{tab:genre-results} in Appendix~\ref{app:genre-res} presents detailed genre-level results on the Test set. CAMeLBERT achieves the best performance in most genres; Poetry is the only exception, where Gemini consistently performs best across all task variants.

\section{Analysis}
\subsection{Word vs. Punctuation Segmentation}
Table~\ref{tab:word-vs-pnx-model-seg} reports Test set performance on word- and punctuation-based sentence boundaries in the punctuated NP and PA settings. Across models, punctuation-based boundaries are substantially easier to detect. CAMeLBERT performs best on both, especially word-based boundaries, where it outperforms all other models by a large margin. Although Gemini-3.1-Pro is the strongest LLM in NP and PA, it remains competitive only on punctuation-based boundaries and struggles with word-based boundaries, particularly in NP.

\begin{table}[t]
\centering
\setlength{\tabcolsep}{4pt}
\begin{tabular}{l|cc|cc}
\toprule
 & \multicolumn{2}{c|}{\textbf{NP}} & \multicolumn{2}{c}{\textbf{PA}} \\
 & \textbf{Pnx} & \textbf{Word} & \textbf{Pnx} & \textbf{Word} \\
\midrule
SaT & 66.6 & 32.1 & 67.7 & 45.8 \\
Gemini-3.1-Pro & 75.5 & 11.0 & 79.4 & 44.5 \\
Flat Orig Deprel & 93.5 & 52.0 & 70.8 & 39.9 \\
CAMeLBERT & \textbf{94.7} & \textbf{60.0} & \textbf{96.3} & \textbf{63.9} \\
\bottomrule
\end{tabular}
\caption{Punctuation- and word-based segmentation results on the Test set of {\AraSEG} in terms of F\textsubscript{1}.}
\label{tab:word-vs-pnx-model-seg}
\end{table}

\begin{table*}[t]
\centering
\setlength{\tabcolsep}{4pt}
\begin{tabular}{l cccc | cccc}
\toprule
& \multicolumn{4}{c}{\textbf{Genre Transferability} (\textbf{$\Delta$F\textsubscript{1}})} & \multicolumn{4}{c}{\textbf{Cross-Genre Benefit} (\textbf{$\Delta$F\textsubscript{1}})}  \\
\midrule
 & \textbf{NoPnx-NP} & \textbf{NoPnx-PA} & \textbf{NP} & \textbf{PA}  & \textbf{NoPnx-NP} & \textbf{NoPnx-PA} & \textbf{NP} & \textbf{PA} \\
\midrule

Children   & -1.7  & -1.9  & -1.9  & -3.7  & -2.5 & -0.2 & -1.1 & \textbf{-0.8} \\
Education  & -10.9 & \textbf{-19.0} & -8.4  & -13.5 & -1.1 & 0 & -0.4 &  0\\
Literature & -1.4  & -1.5  & -2.2  & -4.2 & -0.2 & 0.4 & -1.0 & -0.7 \\
Media      & -0.3  & -2.6  & -5.6  & -6.4 & \textbf{-74.4} & -0.8 & -12.5 &  +0.1 \\
Poetry     & -9.2  & -1.3  & -14.7 & -2.1 & -13.1 &  +1.8 & \textbf{-16.1} &  +1.0 \\
Politics   & -8.0  & -3.9  & -9.6  & -3.3 & +3.3 & -0.9 & -5.1 &  0\\
Religion   & \textbf{-17.3} & -18.4 & \textbf{-30.6} & \textbf{-39.2} & -1.2 & \textbf{-1.9} & -0.2 &  0 \\
Wiki       & -0.6  & +0.6 & -3.3  & -0.2 & -1.3 & -0.4 & -0.4 &  +0.2 \\
\bottomrule
\end{tabular}
\caption{Genre transferability and cross-genre benefit on the Test set. \textbf{Bold} indicates steepest performance decrease.}
\label{tab:logo_results}
\end{table*}

\subsection{Effect of Training Data Size}
\begin{figure}[t!]
    \centering
    \includegraphics[width=\columnwidth]{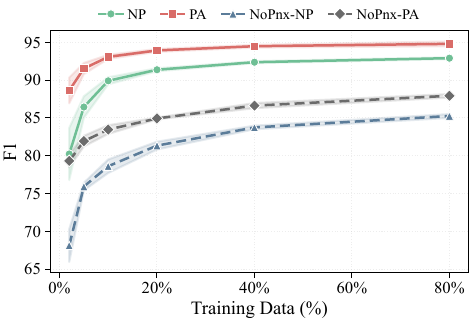}
    \caption{CAMeLBERT Test set performance by training data size.  Shading denotes ±1 standard deviation.}
    \label{fig:data_curve}
\end{figure}

We study the effect of data scale by training on nested 2-80\% training subsets while preserving genre-wise word distributions. Figure~\ref{fig:data_curve} shows CAMeLBERT performance on the Test set averaged over three random seeds. Performance improves with data size, with gains in punctuated settings saturating beyond 40\% of data. In contrast, non-punctuated settings continue improving, reflecting the difficulty of detecting word-based boundaries. Gains are also steeper in NP, especially in low-resource regimes, suggesting that more data compensates for  missing paragraph boundaries.

\subsection{Genre Generalization}
Table~\ref{tab:logo_results} reports genre transferability and cross-genre benefit for CAMeLBERT on the Test set. For genre transferability, we train on all genres except the target genre; for cross-genre benefit, we train only on the target genre. In both cases, $\Delta$F\textsubscript{1} denotes the difference between the corresponding ablated model and full-training-set model.


\paragraph{Genre Transferability} Genre transferability varies substantially across genres. Religion is the least transferable, exhibiting the largest performance drops in NoPnx-NP, NP, and PA, while Education is the most difficult in NoPnx-PA and the second least transferable overall. Poetry ranks third, particularly in NP. In contrast, Wiki, literature, and children remain highly transferable, with minimal variation when excluded from training.


\paragraph{Cross-Genre Benefit} Cross-genre benefit also varies substantially across genres. Media benefits the most from multi-genre training, particularly in NoPnx-NP. Poetry also benefits considerably, especially in NP. In contrast, education, religion, Wiki, and children exhibit only minor gains, indicating limited dependence on other genres.

\begin{table}[t]
\centering
\setlength{\tabcolsep}{4pt}
\begin{tabular}{lc@{\hspace{3pt}}cc@{\hspace{3pt}}cc}
\toprule
& \multicolumn{2}{c}{\textbf{Pnx}}
& \multicolumn{2}{c}{\textbf{Word}}
& \textbf{Total} \\
\midrule
NoPnx-NP & 2,148 & (64\%) & 1,228 & (36\%) & 3,376 \\
NoPnx-PA & 1,611 & (69\%) & 725   & (31\%) & 2,336 \\
NP       & 857   & (56\%) & 676   & (44\%) & 1,533 \\
PA       & 641   & (64\%) & 361   & (36\%) & 1,002 \\

\bottomrule
\end{tabular}
\caption{CAMeLBERT Test set punctuation- and word-based error distribution.}
\label{tab:pnx-word-errors}
\end{table}

\begin{table}[t]
\centering
\setlength{\tabcolsep}{4pt}
\begin{tabular}{lcccc}
\toprule
& \multicolumn{2}{c}{\textbf{NP}}
& \multicolumn{2}{c}{\textbf{PA}} \\
\cmidrule(lr){2-3}\cmidrule(lr){4-5}
\textbf{Pnx}
& \textbf{Prec.}
& \textbf{Err. (\%)}
& \textbf{Prec.}
& \textbf{Err. (\%)} \\
\midrule
, & 79.1 & \textbf{59} & 80.6 & \textbf{70} \\
. & 97.6 & 13 & 99.1 & 9 \\
( & 100 & 0 & 100 & 0 \\
) & 99 & 2 & 99.9 & 1 \\
: & 92.3 & 9 & 97.4 & 5 \\
" & 86.2 & 1 & 100 & 0 \\
? & 99 & 1 & 99 & 1 \\
- & 50 & 1 & 71.4 & 1 \\
.. & 86.3 & 4 & 92.6 & 4 \\
; & 86.8 & 5 & 93.2 & 4 \\
! & 98.6 & 0 & 99.3 & 0 \\
Other & 95.9 & 6 & 97.6 & 5 \\

\bottomrule
\end{tabular}
\caption{CAMeLBERT Test set punctuation-level precision and error distribution in the NP and PA settings.}
\label{tab:np-pa-boundary-precision}
\end{table}

\subsection{Error Analysis}
\paragraph{Punctuation vs. Word Errors} Punctuation-based errors account for most CAMeLBERT errors across all four settings (Table~\ref{tab:pnx-word-errors}), ranging from 56\% in NP to 69\% in NoPnx-PA. Word-based errors remain substantial, reaching 44\% in NP.  Appendix~\ref{app:pnx-word-genre-error-analysis} provides the genre-level breakdown.

\paragraph{Punctuation Errors} Commas dominate punctuation errors (Table~\ref{tab:np-pa-boundary-precision}), consistent with their high ambiguity in {\AraSEG} (\S\ref{sec:pnx-reliability}). Periods contribute the second-largest share despite near-perfect precision, likely reflecting their frequency. Genre-level distributions are in Appendix~\ref{app:pnx-genre-error-analysis}.


\paragraph{Word Errors} We manually categorize genre-stratified 10\% samples of CAMeLBERT's word errors into paragraph boundaries, lexical/discourse ambiguity, verse boundaries, dialogue transitions, and other errors (Appendix~\ref{app:word-error-analysis}). Lexical/discourse ambiguity dominates (56-76\%), while paragraph-boundary errors account for 16-18\% without paragraph information; verse-boundary and dialogue errors comprise most remaining cases.

\section{Segmentation and Dependency Parsing}
We evaluate the impact of sentence segmentation on dependency parsing (Table \ref{tab:parsing-results}) using CamelParser 2.0 on the CAMeLTB Test set. We compare parsing performance using gold segments, SaT segments, CAMeLBERT segments, and no segmentation (Do Nothing), reporting Labeled Attachment Score (LAS), Unlabeled Attachment Score (UAS), and Label Score (LS), standard dependency parsing metrics \cite{nivre-fang-2017-universal}. Sentence segmentation has a substantial effect on parsing performance: LAS drops by nearly 10 points without segmentation. Predicted segmentations partially recover this loss, with CAMeLBERT predictions approaching gold-segmentation performance.



\begin{table}[t!]
\centering
\begin{tabular}{lccc}
\toprule
\textbf{Model} & \textbf{LAS} & \textbf{UAS} & \textbf{LS} \\
\midrule
Do Nothing & 72.6 & 75.3 & 82.5 \\
SaT & 78.0 & 80.4 & 86.0 \\
CAMeLBERT & \textbf{81.5} & \textbf{83.9} & \textbf{87.5} \\
\midrule
Gold & \underline{82.0} & \underline{84.5} & \underline{87.9} \\
\bottomrule
\end{tabular}
\caption{CAMeLTB Test set parsing results. Gold denotes an upper bound. Best results are in \textbf{bold}.
}
\label{tab:parsing-results}
\end{table}

\section{Conclusion and Future Work}
We introduced {\AraSEG}, a multi-genre manually annotated dataset for Arabic sentence segmentation spanning punctuated and unpunctuated text. We benchmarked rule-based methods, dependency parsers, LLMs, and supervised transformers, showing that sentence segmentation remains highly sensitive to genre, punctuation, and document structure. Our results highlight the difficulty of non-punctuated text and word-based boundaries, while demonstrating that supervised transformers outperform alternative approaches. We further show that cross-genre generalization is challenging and that most genres benefit from multi-genre training.

In future work, we plan to extend {\AraSEG} to additional Arabic varieties, investigate more effective segmentation approaches, and study the impact of sentence segmentation on a broader range of downstream Arabic NLP tasks.



\newpage

\section*{Limitations}
While {\AraSEG} covers diverse genres and both punctuated and unpunctuated settings, it remains limited to Modern Standard Arabic and does not include dialectal Arabic varieties. In addition, some genres are represented by fewer sources than others, which may influence transferability analyses. Finally, our experiments focus primarily on encoder-based models and prompting-based LLM evaluation; future work may explore larger instruction-tuned models and segmentation-aware pretraining objectives.

\section*{Ethics Statement}
We used AI writing assistance within the scope of ``Assistance
purely with the language of the paper'' described in the ACL Policy on Publication Ethics.

\section*{Acknowledgments}
This research is in part with support from \href{https://www.google.org/}{Google.org} and the Google Cloud Research Credits program for the Gemini Academic Program.

\bibliography{custom,anthology-1,anthology-2,camel-bib-v3,khalid-custom}

\begin{thebibliography}{64}
\providecommand{\natexlab}[1]{#1}

\bibitem[{Abdelali et~al.(2016)Abdelali, Darwish, Durrani, and Mubarak}]{abdelali-etal-2016-farasa}
Ahmed Abdelali, Kareem Darwish, Nadir Durrani, and Hamdy Mubarak. 2016.
\newblock \href {https://doi.org/10.18653/v1/N16-3003} {{F}arasa: A fast and furious segmenter for {A}rabic}.
\newblock In \emph{Proceedings of the 2016 Conference of the North {A}merican Chapter of the Association for Computational Linguistics: Demonstrations}, pages 11--16, San Diego, California. Association for Computational Linguistics.

\bibitem[{Al-Akkad(1938)}]{akkad:sarah}
Abbas~Mahmoud Al-Akkad. 1938.
\newblock \href {https://www.hindawi.org/books/72707304/} {\emph{Sarah}}.
\newblock Hindawi.

\bibitem[{al~Bukhari(846)}]{bukhari}
Imam~Muhammad al~Bukhari. 846.
\newblock \emph{Sahih al-Bukhari}.
\newblock Dar Ibn Khathir.

\bibitem[{Al-Ghamdi et~al.(2021)Al-Ghamdi, Al-Khalifa, and Al-Salman}]{al-ghamdi-etal-2021-dependency}
Sharefah Al-Ghamdi, Hend Al-Khalifa, and Abdulmalik Al-Salman. 2021.
\newblock \href {https://aclanthology.org/2021.depling-1.1/} {A dependency treebank for classical {A}rabic poetry}.
\newblock In \emph{Proceedings of the Sixth International Conference on Dependency Linguistics (Depling, SyntaxFest 2021)}, pages 1--9, Sofia, Bulgaria. Association for Computational Linguistics.

\bibitem[{Al-Safadi(2005)}]{kashkol}
Bayan Al-Safadi. 2005.
\newblock \emph{Al-Kashkoul: selection of poetry and prose for children ({\footnotesize \<الكشكول: مختارات من الشعر والنثر للأطفال>})}.
\newblock Al-Sa'ih Library ({\footnotesize \<مكتبة السائح>}).

\bibitem[{Alammar et~al.(2025)Alammar, El~Hindi, and Al-Khalifa}]{computation13060151}
Mai Alammar, Khalil El~Hindi, and Hend Al-Khalifa. 2025.
\newblock \href {https://doi.org/10.3390/computation13060151} {English-arabic hybrid semantic text chunking based on fine-tuning bert}.
\newblock \emph{Computation}, 13(6).

\bibitem[{Alfaifi(2015)}]{phdthesis}
A.~Alfaifi. 2015.
\newblock \href {https://doi.org/10.13140/RG.2.2.32081.53608} {\emph{Building the Arabic Learner Corpus and a System for Arabic Error Annotation}}.
\newblock Ph.D. thesis, University of Leeds.

\bibitem[{Alhafni et~al.(2026)Alhafni, Hamed, Eryani, Palfreyman, and Habash}]{alhafni-etal-2026-bilingual}
Bashar Alhafni, Injy Hamed, Fadhl Eryani, David Palfreyman, and Nizar Habash. 2026.
\newblock \href {https://doi.org/10.63317/489vftd6umyh} {A bilingual bimodal benchmark for {A}rabic-{E}nglish {NLP} across grammatical correction, essay scoring, morphological tagging, and speech recognition}.
\newblock In \emph{Proceedings of the Fifteenth Language Resources and Evaluation Conference}, pages 1732--1749, Palma de Mallorca, Spain. ELRA Language Resource Association.

\bibitem[{Alhafni et~al.(2023)Alhafni, Inoue, Khairallah, and Habash}]{alhafni-etal-2023-advancements}
Bashar Alhafni, Go~Inoue, Christian Khairallah, and Nizar Habash. 2023.
\newblock \href {https://doi.org/10.18653/v1/2023.emnlp-main.396} {Advancements in {A}rabic grammatical error detection and correction: An empirical investigation}.
\newblock In \emph{Proceedings of the 2023 Conference on Empirical Methods in Natural Language Processing}, pages 6430--6448, Singapore. Association for Computational Linguistics.

\bibitem[{Alshanqiti et~al.(2022)Alshanqiti, Albouq, Alkhodre, Namoun, and Nabil}]{app122010559}
Abdullah~M. Alshanqiti, Sami Albouq, Ahmad~B. Alkhodre, Abdallah Namoun, and Emad Nabil. 2022.
\newblock \href {https://doi.org/10.3390/app122010559} {Employing a multilingual transformer model for segmenting unpunctuated arabic text}.
\newblock \emph{Applied Sciences}, 12(20).

\bibitem[{Altammami et~al.(2019)Altammami, Atwell, and Alsalka}]{Altammami:2019:Arabic}
Shatha Altammami, Eric Atwell, and Ammar Alsalka. 2019.
\newblock The arabic--english parallel corpus of authentic hadith.
\newblock \emph{International Journal on Islamic Applications in Computer Science And Technology-IJASAT}.

\bibitem[{Ameur et~al.(2008)Ameur, Hassan, and Waleed}]{touir2008semantic}
Touir Ameur, Mathkour Hassan, and Al-Sanea Waleed. 2008.
\newblock \href {https://doi.org/10.3923/itj.2008.1009.1015} {Semantic-based segmentation of arabic texts}.
\newblock \emph{Information Technology Journal}, 7.

\bibitem[{Anwar et~al.(2026)Anwar, Freihat, Ibrahim, Awad, Sadallah, Gosal, Ramakrishnan, Chandran, Mishra, Joshi, Frikha, Goffinet, Maiti, Filali, AlBarri, Ghosh, Pal, Mullah, Shukla, siddiki, Kamboj, Pandit, Sahu, Elbadawy, Mohamed, Chamma, Dufraisse, Bounhar, Bouch, Abdine, Shang, Koto, Wang, Xie, Mekky, Elbadry, Ahmad, Ahsan, Herraoui, Orel, Iqbal, Elzeky, Abassy, Elozeiri, Eletter, Atif, Mukhituly, Li, Han, Singh, Quraishi, Sengupta, Murray, Sheinin, Hestness, Vassilieva, Ren, Liu, Vazirgiannis, and Nakov}]{anwar2026jais2familyarabiccentric}
Mohamed Anwar, Abed~Alhakim Freihat, George Ibrahim, Mostafa Awad, Abdelrahman Sadallah, Gurpreet Gosal, Gokulakrishnan Ramakrishnan, Sarath Chandran, Biswajit Mishra, Rituraj Joshi, Ahmed Frikha, Etienne Goffinet, Abhishek Maiti, Ali~El Filali, Sarah AlBarri, Samujjwal Ghosh, Rahul Pal, Parvez Mullah, Awantika Shukla, and 41 others. 2026.
\newblock \href {https://arxiv.org/abs/2608.13580} {Jais 2: A family of arabic-centric open large language models}.
\newblock \emph{Preprint}, arXiv:2608.13580.

\bibitem[{Beeferman et~al.(1999)Beeferman, Berger, and Lafferty}]{Pk}
Doug Beeferman, Adam Berger, and John Lafferty. 1999.
\newblock \href {https://doi.org/10.1023/A:1007506220214} {Statistical models for text segmentation}.
\newblock \emph{Machine Learning}, 34(1):177--210.

\bibitem[{Bilitski et~al.(2026)Bilitski, Shechter, Jamtsho, Marciano, Bajetta, Sunden, Drori, Hashiloni, Zwebner, Shina, Almogi, Wangchuk, and Bar}]{bilitski-etal-2026-automatic}
Guy Bilitski, Lev Shechter, Sonam Jamtsho, Nir Marciano, Nicola Bajetta, Rebecca Sunden, Omri Drori, Kai~Golan Hashiloni, Orr Zwebner, Asaf Shina, Orna Almogi, Dorji Wangchuk, and Kfir Bar. 2026.
\newblock \href {https://doi.org/10.63317/2iyfjjv9boc6} {Automatic segmentation of classical tibetan texts into autochthonous and allochthonous regions}.
\newblock In \emph{Proceedings of the Fifteenth Language Resources and Evaluation Conference (LREC 2026)}, pages 1017--1030, Palma, Mallorca, Spain. European Language Resources Association (ELRA).

\bibitem[{Bouamor et~al.(2018)Bouamor, Habash, Salameh, Zaghouani, Rambow, Abdulrahim, Obeid, Khalifa, Eryani, Erdmann, and Oflazer}]{bouamor-etal-2018-madar}
Houda Bouamor, Nizar Habash, Mohammad Salameh, Wajdi Zaghouani, Owen Rambow, Dana Abdulrahim, Ossama Obeid, Salam Khalifa, Fadhl Eryani, Alexander Erdmann, and Kemal Oflazer. 2018.
\newblock \href {https://aclanthology.org/L18-1535/} {The {MADAR} {A}rabic dialect corpus and lexicon}.
\newblock In \emph{Proceedings of the Eleventh International Conference on Language Resources and Evaluation ({LREC} 2018)}, Miyazaki, Japan. European Language Resources Association (ELRA).

\bibitem[{Chirkunov et~al.(2026)Chirkunov, Samih, Freihat, and Aldarmaki}]{chirkunov-etal-2026-linear}
Kirill Chirkunov, Younes Samih, Abed~Alhakim Freihat, and Hanan Aldarmaki. 2026.
\newblock \href {https://doi.org/10.18653/v1/2026.findings-acl.1740} {Linear semantic segmentation for low-resource spoken dialects}.
\newblock In \emph{Findings of the {A}ssociation for {C}omputational {L}inguistics: {ACL} 2026}, pages 34844--34861, San Diego, California, United States. Association for Computational Linguistics.

\bibitem[{Comanici et~al.(2025)Comanici, Bieber, Schaekermann, Pasupat, Sachdeva, Dhillon, Blistein, Ram, Zhang, Rosen, Marris, Petulla, Gaffney, Aharoni, Lintz, Pais, Jacobsson, Szpektor, Jiang, Haridasan, Omran, Saunshi, Bahri, Mishra, Chu, Boyd, Hekman, Parisi, Zhang, Kawintiranon, Bedrax-Weiss, Wang, Xu, Purkiss, Mendlovic, Deutel, Nguyen, Langley, Korn, Rossazza, Ramé, Waghmare, Miller, Byrd, Sheshan, Hadsell, Bhardwaj, Janus, Rissa, Horgan, Abdagic, Belenki, Allingham, Singh, Guidroz, Srinivasan, Schmit, Chiafullo, Elisseeff, Jha, Kolhar, Berrada, Ding, Si, Mallick, Och, Erell, Ni, Latkar, Yang, Sirkovic, Feng, Leland, Hornung, Wu, Blundell, Alvari, Huang, Yip, Deur, Liu, Surita, Duque, Damen, Jia, Guez, Mircea, Sinha, Magni, Stradomski, Marian, Galić, Chen, Husain, Singhal, Grewe, Aubet, Song, Blanco, Rechis, Ho, Munoz, Zheng, Hamrick, Mather, Taitelbaum, Rutherford, Lei, Chen, Shukla, Moreira, Doi, Isik, Shabat, Rogozińska, Kolipaka, Chang, Vušak, Venkatachary, Noghabi, Bharti, Jun, Zaks, Green,
  Challagundla, Wong, Mohammad, Hirsch, Cheng, Naim, Proleev, Vincent, Singh, Krikun, Krishnan, Ghahramani, Atias, Aggarwal, Kirov, Vytiniotis, Koh, Chronopoulou, Dogra, Ion, Tyen, Lee, Weissenberger, Strohman, Balakrishna, Rae, Velic, de~Liedekerke, Elyada, Yuan, Liu, Shani, Kishchenko, Alessio, Li, Song, Kwei, Jankowski, Pappu, Namiki, Ma, Tripuraneni, Cherry, Ikonomidis, Ling, Ji, Westberg, Wright, Yu, Parkinson, Ramaswamy, Connor, Yeganeh, Grover, Kenwright, Litchev, Apps, Tomala, Halim, Castro-Ros, Li, Boral, Sho, Yarom, Malmi, Klinghoffer, Lin, Ansell, S, Zhao, Zuo, Santoro, Cheng, Demmessie, Liu, Brichtova, Culp, Braun, Graur, Ng, Mehta, Phillips, Sundberg, Godbole, Liu, Katariya, Rim, Seyedhosseini, Ammirati, Valfridsson, Malihi, Knight, Toor, Lampe, Ittycheriah, Chiang, Yeung, Fréchette, Rao, Wang, Srivastava, Zhang, Rhodes, Brand, Weesner, Figotin, Gimeno, Fellinger, Marcenac, Leal, Marcus, Cotruta, Cabrera, Luo, Garrette, Axelrod, Baltateanu, Barker, Chen, Toma, Ingram, Riesa, Kulkarni, Zhang,
  Liu, Wang, Polacek, Wu, Hui, Reyes, Su, Barnes, Malhi, Siddiqui, Feng, Damaschin, Pighin, Steiner, Yang, Boppana, Ivanov, Kandoor, Shah, Mujika, Huang, Choquette-Choo, Patel, Yu, Creswell, Jerry, Liu, Barros, Razeghi, Roy, Culliton, Xiong, Pan, Strohmann, Powell, Seal, DeCarlo, Shyam, Katircioglu, Wang, Hardin, Odisho, Broder, Chang, Nair, Shtefan, O'Brien, Agarwal, Potluri, Goyal, Jhindal, Thakur, Stuken, Lyon, Toutanova, Feng, Wu, Horn, Wang, Cullum, Taubman, Shrivastava, Shi, Tomlinson, Patel, Tu, Oflazer, Pongetti, Yang, Taïga, Perot, Pierse, Han, Drori, Iturrate, Chakrabarti, Yeung, Dopson, ting Chen, Kulshreshtha, Guo, Pham, Schuster, Chen, Polozov, Xing, Zhou, Kacham, Kukliansky, Miech, Yaroshenko, Chi, Douglas, Fei, Blondel, Myla, Madmoni, Wu, Keysers, Kjems, Albuquerque, Yu, D'sa, Plantan, Ionescu, Elias, Gupta, Vuyyuru, Alcober, Zhou, Ji, Hartmann, Puttagunta, Song, Amid, Stefanoiu, Lee, Pucciarelli, Wang, Raul, Petrov, Tian, Anklin, Nti, Gomes, Schumacher, Vesom, Panagopoulos, Bousmalis, Andor,
  Jacob, Zhang, Rosgen, Kecman, Tung, Belias, Goodman, Covington, Wieder, Saxena, Davoodi, Huang, Maddineni, Roulet, Campbell-Ajala, Sessa, Xintian, Wu, Lai, Collins, Haig, Sakenas, Xu, Giustina, Shafey, Charoenpanit, Garg, Ainslie, Severson, Arenas, Pathak, Rajayogam, Feng, Bakker, Li, Wichers, Rogers, Geng, Li, Jagerman, Jia, Olmert, Sharon, Mauger, Mariserla, Ma, Mohabey, Kim, Andreev, Pollom, Love, Jain, Agrawal, Schroecker, Fortin, Warmuth, Liu, Leach, Blok, Girirajan, Aharoni, Uria, Sozanschi, Goldberg, Ionita, Ribeiro, Zlocha, Birodkar, Lachgar, Yuan, Choudhury, Ginsberg, Zheng, Dibb, Graves, Lokhande, Rasskin, Muraru, Quick, Tata, Sermanet, Chawla, Karo, Wang, Zhang, Keller, Dragan, Su, Chou, Liu, Tao, Prabhakara, Wilson, Liu, Wang, Evans, Du, Castaño, Prasad, Mahdy, Gerlach, Reid, Kahn, Zait, Pillai, Ulrich, Wang, Wassenberg, Farkash, Yalasangi, Wang, Bauza, Bucher, Liu, Yan, Leung, Sindhwani, Barnes, Singh, Jurin, Chang, Bhumihar, Eiger, Citovsky, Withbroe, Li, Xue, Santo, Stoyanov, Raimond, Zheng,
  Gao, Listík, Kwasiborski, Saputro, Ozturel, Mallya, Majmundar, West, Caron, Wei, Castrejon, Vikram, Ramachandran, Dhawan, Park, Smoot, van~den Driessche, Blau, Malik, Liang, Hirsch, dos Santos, Weinstein, van~den Oord, Lall, FitzGerald, Jiang, Yang, Webster, Elqursh, Pope, Rotival, Raposo, Zhu, Dean, Alabed, Tran, Gupta, Gleicher, Austin, Rosseel, Umekar, Das, Sun, Chen, Misiunas, Zhou, Di, Loo, Newlan, Li, Ramasesh, Xu, Chen, Gandhe, Soricut, Gupta, Hu, El-Sayed, Garcia, Brusilovsky, Chen, Bolt, Huang, Gurney, Zhang, Pritzel, Wilkiewicz, Seybold, Shamanna, Fischer, Dean, Gill, Mcilroy, Bhowmick, Selier, Yang, Cheng, Magay, Tan, Varma, Walder, Kocisky, Nakashima, Natsev, Kwong, Gog, Zhang, Dieleman, Jimma, Ryabtsev, Brahma, Steiner, Du, Žužul, Žanić, Raghavachari, Gierke, Zheng, Petrova, Dauphin, Liu, Kessler, Hand, Duvarney, Kim, Lee, Hussenot, Hui, Smith, Jain, Xia, Tomar, Amiri, Phan, Fuchs, Weyand, Tomasev, Cordell, Liu, Mallinson, Joshi, Crawford, Suggala, Chien, Fernando, Sanchez-Vargas,
  Williams, Crone, Luo, Karpov, Shan, Thurk, Strudel, Voigtlaender, Patil, Dozat, Khodaei, Singla, Ambroszczyk, Wu, Chang, Roark, Hegde, Ding, Filos, Wu, Pinto, Liu, Khanna, Pandey, Mcloughlin, Li, Haves, Zhou, Buchatskaya, Leal, de~Boursac, Akazawa, Anderson, Chen, Somandepalli, Liang, Goenka, Winkler, Grushetsky, Ding, Smith, Ye, Pont-Tuset, Li, Li, Golany, Wegner, Jiang, Barak, Shangguan, Vértes, Wong, Bornschein, Tudor, Bevilacqua, Schaul, Rawat, Zhao, Axiotis, Meng, McLean, Lai, Beattie, Kushman, Liu, Kutzman, Lang, Ye, Netrapalli, Mishra, Khan, Goel, Willoughby, Tian, Zhuang, Chen, Tsai, Kementsietsidis, Khare, Keeling, Xu, Waters, Altché, Popat, Mittal, Saxton, Badawy, Mathieu, Zheng, Zhou, Ranka, Shin, Duan, Salimans, Mihailescu, Shaham, Chang, Assael, Dikkala, Izzard, Cohen-Addad, Graves, Feinberg, Chung, Strouse, Karmon, Sharifzadeh, Ashwood, Pham, Blanton, Vasiloff, Barber, Geller, Zhou, Zubach, Huang, Zhang, Gupta, Young, Proskurnia, Votel, Gabeur, Barcik, Tripathi, Yu, Yan, Changpinyo,
  Pavetić, Coyle, Fujii, Mendez, Zhou, Rajamani, Hechtman, Cao, Juan, Tan, Dalibard, Du, Clay, Yao, Jia, Vijaykumar, Zhou, Bai, Hung, Pecht, Todorov, Khadke, Gupta, Lahoti, Autef, Duddu, Lee-Thorp, Bykovsky, Misiunas, Flennerhag, Thangaraj, McGiffin, Nado, Kunesch, Noever, Hertz, Liang, Stone, Palmer, Daruki, Pramanik, Põder, Kyker, Khan, Sluzhaev, Ritter, Ruderman, Zhou, Nagpal, Vodrahalli, Necula, Barham, Pavlick, Hartford, Shafran, Zhao, Mikuła, Eccles, Shimokawa, Garg, Vilnis, Chen, Shumailov, Lee, Abdelhamed, Xie, Cohen, Hlavnova, Malkin, Sitawarin, Lottes, Coquinot, Yu, Kumar, Zhang, Mahendru, Ahmed, Martens, Chen, Boag, Peng, Devin, Klimovskiy, Phuong, Vainstein, Xie, Ramabhadran, Howard, Yu, Goswami, Cui, Shleifer, Pinto, Yeh, Yang, Javanmardi, Ethier, Lee, Orbay, Kotecha, Bromberg, Shaw, Thornton, Rosenthal, Gu, Thomas, Gemp, Ayyar, Ushio, Selvan, Wee, Liu, Majzoubi, Yu, Abernethy, Liechty, Pan, Nguyen, Qiong, Hu, Perrin, Arora, Pitler, Wang, Shivakumar, Prost, Limonchik, Wang, Gao, Cour, Buch,
  Gui, Ivanova, Neubeck, Chan, Kim, Chen, Goyal, Chung, Liu, Su, Petrushkina, Shen, Joulin, Xu, Lin, Kulizhskaya, Chelba, Vasudevan, Collins, Bashlovkina, Lu, Fritz, Park, Zhou, Su, Tanburn, Sushkov, Rasquinha, Li, Prendki, Li, LV, Sharma, Fitoussi, Huang, Dai, Dao, Burrows, Prior, Qin, Pundak, Sjoesund, Khurshudov, Zhu, Webson, Kemp, Tan, Agrawal, Sargsyan, Cheng, Stephan, Kwiatkowski, Reid, Byravan, Michaely, Heess, Zhou, Goenka, Carpenter, Levskaya, Wang, Roberts, Leblond, Chikkerur, Ginzburg, Chang, Riachi, Chuqiao, Xu, Borsos, Pliskin, Pawar, Lustman, Kirkwood, Anand, Chaudhary, Kalb, Milan, Augenstein, Goldie, Prince, Raman, Sun, Xia, Cohen, Huo, Camp, Ellis, Zilka, Torres, Patel, Arora, Chan, Adler, Ayoub, Liang, Jamil, Jiang, Baumgartner, Sun, Karov, Akulov, Zheng, Cai, Fantacci, Rubin, Acha, Wang, D'Souza, Sathyanarayana, Dai, Rowe, Simanovsky, Goldman, Kuang, Pan, Rosenberg, Rojas-Esponda, Dutta, Zeng, Jurenka, Farquhar, Bansal, Iqbal, Roelofs, Joung, Beak, Ryu, Poplin, Wu, Alayrac, Buthpitiya,
  Ronneberger, Habtegebriel, Li, Cavallaro, Wei, Bensky, Denk, Ganapathy, Stanway, Joshi, Bertolini, Lo, Ma, Charles, Sampemane, Sahni, Chen, Askham, Gaddy, Young, Tan, Eyal, Bražinskas, Zhong, Wu, Epstein, Bailey, Hard, Lee, Goldshtein, Ruiz, Badawi, Lochbrunner, Kearns, Brown, Pardo, Weber, Yang, Jiang, Akin, Fu, Wainwright, Zou, Gaba, Manzagol, Kan, Song, Zainullina, Lin, Ko, Deshmukh, Jindal, Svensson, Tyam, Zhao, Kaeser-Chen, Baird, Moradi, Hall, Guo, Tsang, Liang, Pereira, Ganesh, Korotkov, Adamek, Thiagarajan, Tran, Chen, Tar, Jain, Dasgupta, Bilal, Reitter, Zhao, Vezzani, Gehman, Mehta, Beltrone, Dotiwalla, Guadarrama, Abbas, Karp, Georgiev, Ferng, Brockschmidt, Peng, Hirnschall, Verma, Bi, Xiao, Dabush, Xu, Wallis, Parker, Wang, Xu, Safarli, Tewari, Zhang, Kim, Gesmundo, Thomas, Levi, Chowdhury, Rao, Garst, Conway-Rahman, Ran, McKinney, Xiao, Yu, Agrawal, Stjerngren, Ionescu, Chen, Sharma, Chiu, Liu, Franko, Sanford, Cai, Michel, Ganapathy, Labanowski, Garrett, Vargas, Sun, Gale, Buschmann,
  Desjardins, Ghelani, Jain, Verma, Asawaroengchai, Eisenschlos, Harlalka, Kazawa, Metzler, Howland, Jian, Ades, Shah, Gangwani, Lee, Ring, Hernandez, Reich, Sinha, Sathe, Kovac, Gill, Kannan, D'olimpio, Sevenich, Whang, Kim, Sim, Chen, Zhang, Lall, Matias, Jia, Friesen, Nasso, Thapliyal, Perozzi, Yu, Shekhawat, Huda, Grabowski, Wang, Sreevatsa, Dib, Hassen, Schuh, Milutinovic, Welty, Quinn, Shah, Wang, Barth-Maron, Frye, Axelsson, Zhu, Ma, Giannoumis, Sedghi, Ye, Luan, Aydin, Chandra, Sampathkumar, Huang, Lavrenko, Eleryan, Hong, Hansen, Carthy, Samanta, Ćevid, Wang, Li, Voznesensky, Hoffman, Terzis, Sehwag, Fidel, He, Cai, He, Feng, Nikoltchev, Phatale, Chase, Lawton, Zhang, Ouyang, Tragut, Manshadi, Narayanan, Shen, Gao, Bolukbasi, Roy, Li, Golovin, Panait, Qin, Han, Anthony, Kudugunta, Patraucean, Ray, Chen, Yang, Bhatia, Talluri, Morris, Ražnatović, Brownfield, An, Peng, Kane, Zheng, Duduta, Kessinger, Noraky, Liu, Rong, Veličković, Rush, Goldin, Wei, Garlapati, Pantofaru, Kwon, Ni, Noland, Trapani,
  Beaufays, Roy, Chow, Turker, Cideron, Mei, Clark, Dou, Bošnjak, Leith, Du, Yazdanbakhsh, Nasr, Kwak, Sheth, Kaskasoli, Anand, Lakshminarayanan, Jerome, Bieber, Chu, Senges, Shen, Sridhar, Ndebele, Beyret, Mohamed, Chen, Freitag, Guo, Liu, Roit, Chen, Yan, Stone, Co-Reyes, Cole, Scellato, Azizi, Hashemi, Jin, Iyer, Valentine, György, Ahuja, Diaz, Lee, Clement, Kong, Garmon, Watts, Bhatia, Gupta, Miecnikowski, Vallet, Taly, Loper, Joshi, Atwood, Chick, Collier, Iliopoulos, Trostle, Gunel, Leal-Cavazos, Hrafnkelsson, Guzman, Ju, Forbes, Emond, Chauhan, Caine, Xiao, Zeng, Moufarek, Murphy, Meng, Gupta, Riedel, Das, Lawal, Narayan, Sosea, Swirhun, Friso, Neyshabur, Lu, Girgin, Wunder, Yvinec, Pyne, Carbune, Rijhwani, Guo, Doshi, Briukhov, Bain, Hitron, Wang, Gupta, Chen, Du, Zhang, Shah, Akula, Dylla, Kachra, Kuo, Zou, Wang, Xu, Zhu, Snyder, Menon, Firat, Mordatch, Yuan, Ponomareva, Blevins, Moore, Wang, Chen, Scholz, Dwornik, Lin, Li, Antognini, I, Song, Miller, Kalra, Raveret, Akerlund, Wu, Nystrom, Godbole,
  Liu, DeBalsi, Zhao, Liu, Caciularu, Lax, Khandelwal, Langston, Bailey, Lattanzi, Wang, Kovelamudi, Mondal, Guruganesh, Hua, Roval, Wesołowski, Ingale, Halcrow, Sohn, Angermueller, Raad, Stickgold, Lu, Kosik, Xie, Lillicrap, Huang, Zhang, Paulus, Farabet, Wertheim, Wang, Joshi, ling Ko, Wu, Agrawal, Lin, Sheng, Sung, Breland-King, Butterfield, Gawde, Singh, Zhang, Apte, Shetty, Hutter, Li, Salesky, Lebron, Kanerva, Paganini, Nguyen, Vallu, Peter, Velury, Kao, Hoover, Bortsova, Bishop, Jakobovits, Agostini, Agarwal, Liu, Kwong, Tavakkol, Bica, Greve, GP, Marcus, Hou, Duerig, Moroshko, Lacey, Davis, Amelot, Wang, Kim, Strinopoulos, Wan, Lan, Krishnan, Tang, Humphreys, Bai, Shtacher, Machado, Pang, Burke, Liu, Aravamudhan, Song, Hirst, Singh, Jou, Bai, Piccinno, Fu, Alazard, Meiri, Winter, Chen, Zhang, Heitkaemper, Lambert, Lee, Frömmgen, Rogulenko, Nair, Niemczyk, Bulyenov, Xu, Shemtov, Zadimoghaddam, Toropov, Wirth, Dai, Gollapudi, Zheng, Kurakin, Lee, Bullard, Serrano, Balazevic, Li, Schalkwyk, Murphy,
  Zhang, Sequeira, Datta, Agrawal, Sutton, Attaluri, Chiang, Farhan, Thornton, Lin, Choma, Nguyen, Dasgupta, Robinson, Comşa, Riley, Pillai, Mustafa, Golan, Zandieh, Lespiau, Porter, Ross, Rajayogam, Agarwal, Venugopalan, Shahriari, Yan, Xu, Tobin, Dubov, Shi, Recasens, Kovsharov, Borgeaud, Dery, Vasanth, Gribovskaya, Qiu, Mahdieh, Skut, Nielsen, Zheng, Yu, Bostock, Gupta, Archer, Rawles, Davies, Svyatkovskiy, Tsai, Halpern, Reisswig, Wydrowski, Chang, Puigcerver, Taege, Li, Schnider, Li, Dena, Xu, Telang, Shi, Zen, Kastner, Ko, Subramaniam, Kumar, Blois, Dai, Wieting, Lu, Zeldes, Xie, Hauth, Ţifrea, Li, El-Husseini, Abolafia, Zhou, Ding, Ghalebikesabi, Guía, Maksai, Ágoston Weisz, Arik, Sukhanov, Świetlik, Jia, Yu, Wang, Brand, Bloxwich, Kirmani, Chen, Go, Sprechmann, Kannen, Carin, Sandhu, Edkins, Nooteboom, Gupta, Maggiore, Azizi, Pritch, Yin, Gupta, Tarlow, Smith, Ivanov, Babaeizadeh, Goel, Kambala, Chu, Kastelic, Liu, Soltau, Stone, Agrawal, Kim, Soparkar, Tadepalli, Bunyan, Soh, Kannan, Kim, Chen,
  Halumi, Roy, Wang, Sercinoglu, Gibson, Bhatnagar, Sano, von Dincklage, Ren, Mitrevski, Olšák, She, Doersch, Jilei, Wang, Liu, Tan, Yakar, Warkentin, Ramirez, Lebsack, Dillon, Mathews, Cobley, Wu, Chen, Simon, Nath, Sainath, Bendebury, Julian, Mankalale, Ćurko, Zacchello, Brown, Sodhia, Howard, Caelles, Gupta, Evans, Bulanova, Katzen, Goldenberg, Tsitsulin, Stanton, Schillings, Kovalev, Fry, Shah, Lin, Upadhyay, Li, Radpour, Maggioni, Xiong, Haas, Brennan, Kamath, Savinov, Nagrani, Yacovone, Kappedal, Andriopoulos, Lao, Li, Rozhdestvenskiy, Hashimoto, Audibert, Austin, Rodriguez, Ruoss, Honke, Karkhanis, Xiong, Wei, Huang, Leng, Premachandran, Bileschi, Evangelopoulos, Mensink, Pavagadhi, Teplyashin, Chang, Xue, Tanzer, Goldman, Patel, Li, Wiesner, Zheng, Stewart-Binks, Han, Li, Luo, Lenc, Lučić, Xue, Mullins, Guseynov, Chang, Galatzer-Levy, Zhang, Bingham, Hu, Hartman, Ma, Griffith, Irpan, Radebaugh, Yue, Fan, Ungureanu, Sorokin, Teufel, Li, Anil, Paparas, Wang, Lin, Peng, Shum, Petrovic, Brady,
  Nguyen, Macherey, Li, Singh, Yenugula, Iinuma, Chen, Kopparapu, Stern, Dave, Thekkath, Perot, Kumar, Li, Xiao, Bilotti, Bateni, Noble, Lee, Vázquez-Reina, Salazar, Yang, Wang, Gruzewska, Rao, Raghuram, Xu, Ben-David, Mei, Dalmia, Zhang, Liu, Bansal, Pankov, Schwarcz, Burns, Chan, Sanghai, Liang, Liang, He, Stuart, Narayanan, Zhu, Frank, Fatemi, Sabne, Lang, Bhattacharya, Settle, Wang, McMahan, Tacchetti, Soares, Hadian, Cabi, Chung, Putikhin, Li, Chen, Tarango, Michalewski, Kazemi, Masoom, Sheftel, Shivanna, Vadali, Comanescu, Reid, Moore, Neelakantan, Sander, Herzig, Rosenberg, Dehghani, Choi, Fink, Hayes, Ge, Weng, Ho, Karro, Krishna, Thiet, Skerry-Ryan, Eppens, Andreetto, Sarma, Bonacina, Ayan, Nawhal, Shan, Dusenberry, Thakoor, Gubbi, Nguyen, Tsarfaty, Albanie, Mitrović, Gandhi, Chen, Epasto, Stephanov, Jin, Gehman, Amini, Weber, Behbahani, Xu, Allamanis, Chen, Ott, Sha, Jastrzebski, Qi, Greene, Wu, Toki, Vlasic, Shapiro, Kotikalapudi, Shen, Saeki, Xie, Cassirer, Bharadwaj, Kiyono, Bhojanapalli,
  Rosenfeld, Ritter, Mao, Oliveira, Egyed, Bandemer, Parisotto, Kinoshita, Pluto, Maniatis, Li, Guo, Ghiasi, Tarbouriech, Chatterjee, Jin, Katrina, Xu, Palomaki, Arnold, Sewak, Piccinini, Sharma, Albrecht, Purser-haskell, Vaswani, Chen, Wisniewski, Cao, Aslanides, Phu, Sieb, Agubuzu, Zheng, Sohn, Selvi, Andreassen, Subudhi, Eruvbetine, Woodman, Mery, Krause, Ren, Ma, Luo, Chen, Fan, Griffiths, Schuler, Li, Zhang, Sarr, Luo, Patana, Watson, Naboulsi, Collins, Sidhwani, Hoogeboom, Silver, Caveness, Zhao, Rodriguez, Deines, Bai, Griffin, Tagliasacchi, Xue, Babbula, Pang, Ding, Shen, Peake, Crocker, Raghvendra, Swisher, Han, Singh, Wu, Pchelin, Munkhdalai, Alon, Bacon, Robles, Bulian, Johnson, Powell, Ferreira, Li, Benzing, Velimirović, Soyer, Kong, Tony, Nguyên, Yang, Liu, van Amersfoort, Gillick, Sun, Rauschmayr, Zhang, Zhan, Zhou, Frolov, Yang, Vnukov, Rouillard, Li, Mandhane, Fallen, Venkataraman, Hu, Brennan, Lee, Chang, Sundermeyer, Pan, Ke, Tong, Fabrikant, Bono, Gu, Foley, Mao, Delakis, Bhaswar,
  Frostig, Li, Zipori, Hope, Kozlova, Mishra, Djolonga, Schiff, Merey, Briakou, Morgan, Wan, Hassidim, Skerry-Ryan, Sengupta, Jasarevic, Kallakuri, Kunkle, Brennan, Lieber, Mansoor, Walker, Zhang, Xie, Žužić, Chukwuka, Druinsky, Cho, Yao, Naeem, Butt, Kim, Jia, Jordan, Lelkes, Kurzeja, Wang, Zhao, Over, Chakladar, Prasetya, Jha, Ganapathy, Cong, Shroff, Saroufim, Miryoosefi, Hammad, Nasir, Xi, Gao, Maeng, Hora, Cheng, Haghani, Lewenberg, Lu, Matysiak, Raisinghani, Wang, Baugher, Sukthankar, Giang, Schultz, Fiedel, Chen, Lee, Dey, Zheng, Paul, Smith, Ly, Wang, Bansal, Perz, Ricco, Blank, Keshava, Sharma, Chow, Lad, Jalan, Osindero, Swanson, Scott, Ilić, Li, Jonnalagadda, Soudagar, Xiong, Batsaikhan, Jarrett, Kumar, Shah, Lawlor, Waters, Graham, May, Ramos, Lefdal, Cankara, Cano, O'Donoghue, Borovik, Liu, Grimstad, Alnahlawi, Tsihlas, Hudson, Grigorev, Jia, Huang, Igwe, Lebedev, Tang, Krivokon, Garcia, Tan, Jia, Stys, Vashishth, Liang, Venkatraman, Gu, Kementsietsidis, Zhu, Jung, Bai, Hosseini, Ahmed,
  Gupta, Yuan, Ashraf, Nigam, Vasudevan, Awasthi, Gilady, Mariet, Eskander, Li, Hu, Garrido, Schlattner, Zhang, Saxena, Dević, Muralidharan, Murthy, Zhou, Choi, Wongpanich, Wang, Shah, Xu, Huang, Spencer, Chen, Cohan, Wang, Tompson, Wu, Haroun, Li, Huergo, Yang, Yin, Wendt, Bendersky, Chaabouni, Snaider, Ferret, Jindal, Thompson, Xue, Bishop, Phal, Sharma, Sung, Radhakrishnan, Shomrat, Ingle, Vij, Gilmer, Istin, Sobell, Lu, Nottage, Sadigh, Willcock, Zhang, Xu, Brown, Lee, Wang, Zhu, Tay, Kim, Gutierrez, Sharma, Xian, Seo, Cui, Pochernina, Baetu, Jastrzębski, Ly, Elhawaty, Suh, Sezener, Wang, Yuen, Tucker, Cai, Yang, Wang, Muzio, Qian, Yoo, Lockhart, McKee, Guo, Mehrotra, Mendonça, Mehta, Ben, Tekur, Mu, Zhu, Krakovna, Lee, Maschinot, Cevey, Choe, Bai, Srinivasan, Gasaway, Young, Siegler, Holtmann-Rice, Piratla, Baumli, Yogev, Hofer, van Hasselt, Grant, Chervonyi, Silver, Hogue, Agarwal, Wang, Singh, Flynn, Lipschultz, David, Bellot, Yang, Le, Graziano, Olszewska, Hui, Maurya, Parotsidis, Chen, Oguntebi,
  Kelley, Baddepudi, Mauerer, Shaw, Siegman, Yang, Shetty, Roy, Song, Stokowiec, Burnell, Savant, Busa-Fekete, Miao, Ghosh, MacDermed, Lippe, Dektiarev, Behrman, Mentzer, Nguyen, Wei, Verma, Knutsen, Dasari, Yan, Mitrichev, Wang, Shejwalkar, Austin, Sunkara, Potti, Virin, Wright, Liu, Riva, Pot, Kochanski, Le, Balasubramaniam, Dhar, Liao, Bloniarz, Shukla, Cole, Lee, Zhang, Kafle, Vashishtha, Mahmoudieh, Chen, Hoffmann, Srinivasan, Lago, Shalom, Wang, Elabd, Sharma, Oh, Kothawade, Le, Monteiro, Yang, Alarakyia, Geirhos, Mincu, Garnes, Kobayashi, Mariooryad, Krasowiak, Zhixin, Lai, Mourad, Wang, Bu, Aharoni, Chen, Goyal, Zubov, Bapna, Dabir, Kothari, Lamerigts, Cao, Shar, Yew, Kulkarni, Mahaarachchi, Joshi, Zhu, Lichtarge, Zhou, Muckenhirn, Selo, Vinyals, Chen, Brohan, Mehta, Cogan, Wang, Geri, Ko, Chen, Viola, Shivam, Wang, Elish, Popa, Pereira, Liu, Koster, Kim, Zhang, Ebrahimi, Talukdar, Zheng, Poklukar, Mikhalap, Johnson, Vijayakumar, Omernick, Dibb, Dubey, Hu, Suman, Aggarwal, Kornakov, Xia, Lowe,
  Kolganov, Xiao, Nikolaev, Hemingray, Li, Iljazi, Rybiński, Sandhu, Lu, Luong, Jenatton, Govindaraj, Hui, Li, Dulac-Arnold, Park, Wang, Modi, Pouget-Abadie, Greller, Gupta, Berry, Ramachandran, Xie, McCafferty, Wang, Gupta, Lim, Bratanič, Brock, Akolzin, Sproch, Karliner, Kim, Goedeckemeyer, Shazeer, Schmid, Calandriello, Bhatia, Choromanski, Montgomery, Dua, Ramalho, King, Gao, Nguyen, Lindner, Pitta, Johnson, Salama, Ardila, Han, Farnese, Odoom, Wang, Ding, Rink, Smith, Lehri, Cohen, Vats, He, Gopavarapu, Paszke, Patel, Gansbeke, Loher, Castro, Voitovich, von Glehn, George, Niklaus, Eaton-Rosen, Rakićević, Jue, Perel, Zhang, Bahat, Pouget, Xing, Huot, Shenoy, Bos, Coriou, Richter, Noy, Wang, Ontanon, Qin, Makarchuk, Hassabis, Li, Sharma, Venkatesan, Kemaev, Daniel, Huang, Shah, Ponce, Warren, Chen, Faruqui, Wu, Andačić, Payrits, McDuff, Hume, Cao, Tessler, Wang, Wang, Rendulic, Agustsson, Johnson, Lando, Howard, Padmanabhan, Daswani, Banino, Kilgore, Heek, Ji, Caceres, Li, Kassner, Vlaskin, Liu,
  Grills, Hou, Sukkerd, Cheon, Shetty, Markeeva, Stanczyk, Iyer, Gong, Gao, Gopalakrishnan, Blyth, Reynolds, Bhoopchand, Bilenko, Gharibian, Zayats, Faust, Singh, Ma, Jiao, Vijayanarasimhan, Aroyo, Yadav, Chakera, Kakarla, Meshram, Gregor, Botea, Senter, Jia, Kovacs, Sharma, Baur, Kang, He, Zhuo, Kostelac, Laish, Peng, O'Bryan, Kasenberg, Rao, Leurent, Zhang, Stevens, Salazar, Zhang, Lobov, Walker, Porter, Redshaw, Ke, Rao, Lee, Lam, Moffitt, Kim, Qiao, Koo, Dadashi, Song, Sundararajan, Xu, Kawamoto, Zhong, Barbu, Reddy, Verzetti, Li, Papamakarios, Klimczak-Plucińska, Cassin, Kavukcuoglu, Swavely, Vaucher, Zhao, Hemsley, Tschannen, Ge, Menghani, Yu, Ha, He, Wu, Song, Sterneck, Zinke, Calian, Marsden, Ruiz, Hessel, Gueta, Lee, Farris, Gupta, Li, Saleh, Misra, Xiao, Mendolicchio, Buttimore, Krayvanova, Nayakanti, Wiethoff, Pande, Mirhoseini, Lao, Liu, Hua, Chen, Malkov, Kalashnikov, Gupta, Audhkhasi, Zhai, Kopalle, Jain, Ofek, Meyer, Baatarsukh, Strejček, Qian, Freedman, Figueira, Sokolik, Bachem, Lin,
  Kharrat, Hidey, Xu, Duan, Li, Ersoy, Everett, Cen, Santamaria-Fernandez, Taubenfeld, Mackinnon, Deng, Zablotskaia, Viswanadha, Goel, Yates, Deng, Choy, Chen, Sinha, Mossin, Wang, Szlam, Hao, Rubenstein, Toksoz-Exley, Aperghis, Zhong, Ahn, Isard, Lacombe, Luisier, Anastasiou, Kalley, Prabhu, Dunleavy, Bijwadia, Mao-Jones, Chen, Pasumarthi, Wood, Dostmohamed, Hurley, Simsa, Parrish, Pajarskas, Harvey, Skopek, Kochinski, Rey, Rieser, Zhou, Lee, Acharya, Li, Jiang, Zhang, Gipson, Mahintorabi, Gelmi, Khajehnouri, Yeh, Lee, Matthey, Baker, Pham, Fu, Pak, Gupta, Vasconcelos, Sadovsky, Walker, Hsiao, Zochbauer, Marzoca, Velan, Zeng, Baechler, Driess, Jain, Huang, Tao, Maggs, Levine, Schneider, Gemzer, Petit, Han, Fisher, Zelle, Biles, Ie, Fadeeva, Liu, Franco, Collister, Zhang, Wang, Zhao, Kieliger, Shuster, Zhu, Gong, Chan, Sun, Basu, Zimmermann, Hayes, Bapna, Snoek, Yang, Datta, Abdallah, Kilgour, Li, Mah, Jun, Rivière, Karmarkar, Spalink, Huang, Gonzalez, Tran, Nowak, Palowitch, Chadwick, Talius, Mehta, Sellam,
  Fränken, Nicosia, He, Kini, Amos, Basu, Jobe, Shaw, Xu, Evans, Ikeda, Yan, Jin, Wang, Yadav, Labzovsky, Sampath, Ma, Schumann, Siddhant, Shah, Youssef, Agarwal, Dabney, Tonioni, Ambar, Li, Guyon, Li, Soergel, Fang, Karadzhov, Udrescu, Trinh, Raunak, Noury, Guo, Gupta, Finkelstein, Petek, Liang, Billock, Sun, Wood, Song, Yu, Matejovicova, Cohen, Andra, D'Ambrosio, Deng, Nallatamby, Songhori, Dangovski, Lampinen, Botadra, Hillier, Cao, Baddi, Kuncoro, Yoshino, Bhagatwala, Ranzato, Schaeffer, Liu, Ye, Sarvana, Nham, Kuang, Gao, Baek, Mittal, Wahid, Gergely, Ni, Feldman, Muir, Lamblin, Macherey, Dyer, Kilpatrick, Campos, Bhutani, Fort, Ahmad, Severyn, Chatziprimou, Ferludin, Dimarco, Kusupati, Heyward, Bahir, Villela, Millican, Marcus, Bahargam, Unlu, Roth, Wei, Gopal, Ghoshal, Lee, Lin, Lees, Lee, Hosseini, Fan, Neel, Wu, Altun, Cai, Piqueras, Woodward, Bissacco, Haykal, Bordbar, Sundaram, Hodkinson, Toyama, Polovets, Myers, Sinha, Levinboim, Krishnakumar, Chhaparia, Sholokhova, Gundavarapu, Jawahar, Qureshi,
  Hu, Momchev, Rahtz, Wu, S, Dhamdhere, Guo, Gupta, Eslami, Schain, Blokzijl, Welling, Orr, Bolelli, Perez-Nieves, Sirotenko, Prasad, Kar, Pigem, Terzi, Weisz, Ghosh, Mavalankar, Madeka, Daugaard, Adam, Shah, Berman, Tran, Baker, Andrejczuk, Chole, Raboshchuk, Mirzazadeh, Kagohara, Wu, Schallhart, Orlando, Wang, Rrustemi, Xiong, Liu, Vezer, Ramsden, yiin Chang, Mudgal, Li, Vieillard, Hoshen, Ahmad, Slone, Hua, Potikha, Rossini, Stritar, Prakash, Wang, Dong, Nazari, Nehoran, Tekelioglu, Li, Badola, Funkhouser, Li, Yerram, Ganeshan, Formoso, Langner, Shi, Li, Yamamori, Panda, Saade, Scarpati, Breaux, Carey, Zhou, Hsieh, Bridgers, Butryna, Gupta, Tulsyan, Woo, Eltyshev, Grathwohl, Parks, Benjamin, Panigrahy, Dodhia, Freitas, Sauer, Song, Alet, Tolins, Paduraru, Zhou, Albert, Zhang, Shu, Bansal, Nguyen, Globerson, Xiao, Manyika, Hennigan, Rong, Matak, Bakalov, Sharma, Sinopalnikov, Pierson, Roller, Brown, Gao, Fukuzawa, Ghafouri, Vassigh, Barr, Wang, Korsun, Jayaram, Ren, Zaman, Khan, Lunts, Deutsch, Uthus, Katz,
  Samsikova, Khalifa, Sethi, Sun, Tang, Alon, Luo, Yu, Nayyar, Petrini, Truong, Hellendoorn, Chinaev, Alberti, Wang, Hu, Mirrokni, Balashankar, Aharon, Mehta, Iscen, Kready, Manning, Mohananey, Chen, Tripathi, Wu, Petrovski, Hwang, Baeuml, Chandrakaladharan, Liu, Coaguila, Chen, Ma, Tafti, Tatineni, Spitz, Ye, Vicol, Rosca, Puigdomènech, Yahav, Ghemawat, Lin, Kirk, Nabulsi, Brin, Bohnet, Caluwaerts, Veerubhotla, Zheng, Dai, Petrov, Xu, Mehran, Xu, Zintgraf, Choi, Hombaiah, Thoppilan, Reddi, Lew, Li, Webster, Sawhney, Lamprou, Shakeri, Lunayach, Chen, Bagri, Salcianu, Chen, Donchev, Magister, Nørly, Rodrigues, Izo, Noga, Zou, Köppe, Zhou, Lee, Long, Eisenbud, Chen, Schenck, To, Zhong, Taropa, Truong, Levy, Martins, Zhang, Semturs, Zhang, Yakubovich, Moreno, McConnaughey, Lu, Redmond, Weerts, Bitton, Refice, Lacasse, Conmy, Tallec, Odell, Forbes-Pollard, Socala, Hoech, Kohli, Walton, Wang, Sazanovich, Zhu, Kapishnikov, Galt, Denton, Murdoch, Sikora, Mohamed, Wei, First, McConnell, Cobo, Qin, Avrahami, Balle,
  Watanabe, Louis, Kraft, Ariafar, Gu, Rives, Yoon, Rusu, Cobon-Kerr, Hahn, Luo, Yuvein, Zhu, Ahuja, Benenson, Kaufman, Yu, Hightower, Zhang, Ni, Hendricks, Wang, Yona, Jain, Barrio, Bhupatiraju, Velusamy, Dafoe, Riedel, Thomas, Yuan, Bellaiche, Panthaplackel, Kloboves, Jauhari, Akbulut, Davchev, Gladchenko, Madras, Chuklin, Hill, Yuan, Madhavan, Leonhard, Scandinaro, Chen, Niu, Douillard, Damoc, Onoe, Pedregosa, Bertsch, Leichner, Pagadora, Malmaud, Ponda, Twigg, Duzhyi, Shen, Wang, Garg, Chen, Evci, Lee, Liu, Kojima, Yamaguchi, Rajendran, Piergiovanni, Rajendran, Fornoni, Ibagon, Ragan, Khan, Blitzer, Bunner, Sun, Kosakai, Lundberg, Elue, Guu, Park, Park, Narayanaswamy, Wu, Mudigonda, Cohn, Mu, Kumar, Graesser, Zhang, Killam, Zhuang, Giménez, Jishi, Ley-Wild, Zhai, Osawa, Cedillo, Liu, Upadhyay, Sieniek, Sharma, Paine, Angelova, Addepalli, Parada, Majumder, Lamp, Kumar, Deng, Myaskovsky, Sabolić, Dudek, York, de~Chaumont~Quitry, Nie, Cattle, Gunjan, Piot, Khawaja, Bang, Wang, Khodadadeh, R, Rawlani,
  Powell, Lee, Griesser, Oh, Magalhaes, Li, Tokumine, Vogel, Hsu, BC, Jindal, Cohen, Yang, Yuan, de~Cesare, Bruguier, Xu, Roy, Jacovi, Belov, Arya, Meadowlark, Cohen-Ganor, Ye, Morris-Suzuki, Banzal, Song, Ponnuramu, Zhang, Scrivener, Zaiem, Rochman, Han, Ghazi, Lee, Drath, Suo, Girgis, Shenoy, Nguyen, Eck, Gupta, Yan, Carreira, Gulati, Sang, Mirylenka, Cooney, Chou, Ling, Fan, Coleman, Tubone, Kumar, Baldridge, Hernandez-Campos, Lazaridou, Besley, Yona, Bulut, Wellens, Pierigiovanni, George, Green, Han, Tao, Clark, You, Abdolmaleki, Fu, Chen, Chaugule, Chandorkar, Rahman, Thompson, Koanantakool, Bernico, Ren, Vlasov, Vassilvitskii, Kula, Liang, Kim, Huang, Ye, Lepikhin, and Helmholz}]{gemini}
Gheorghe Comanici, Eric Bieber, Mike Schaekermann, Ice Pasupat, Noveen Sachdeva, Inderjit Dhillon, Marcel Blistein, Ori Ram, Dan Zhang, Evan Rosen, Luke Marris, Sam Petulla, Colin Gaffney, Asaf Aharoni, Nathan Lintz, Tiago~Cardal Pais, Henrik Jacobsson, Idan Szpektor, Nan-Jiang Jiang, and 3416 others. 2025.
\newblock \href {https://arxiv.org/abs/2507.06261} {Gemini 2.5: Pushing the frontier with advanced reasoning, multimodality, long context, and next generation agentic capabilities}.
\newblock \emph{Preprint}, arXiv:2507.06261.

\bibitem[{Dozat and Manning(2016)}]{DBLP:journals/corr/DozatM16}
Timothy Dozat and Christopher~D. Manning. 2016.
\newblock \href {https://arxiv.org/abs/1611.01734} {Deep biaffine attention for neural dependency parsing}.
\newblock \emph{CoRR}, abs/1611.01734.

\bibitem[{Dukes et~al.(2013)Dukes, Atwell, and Habash}]{Dukes:2013:supervised}
Kais Dukes, Eric Atwell, and Nizar Habash. 2013.
\newblock Supervised collaboration for syntactic annotation of quranic arabic.
\newblock \emph{Language resources and evaluation}, 47(1):33--62.

\bibitem[{Eck and Hori(2005)}]{eck-hori-2005-overview}
Matthias Eck and Chiori Hori. 2005.
\newblock \href {https://aclanthology.org/2005.iwslt-1.1/} {Overview of the {IWSLT} 2005 evaluation campaign}.
\newblock In \emph{Proceedings of the Second International Workshop on Spoken Language Translation}, Pittsburgh, Pennsylvania, USA.

\bibitem[{El-Haj and Ezzini(2024)}]{el-haj-ezzini-2024-multilingual}
Mo~El-Haj and Saad Ezzini. 2024.
\newblock \href {https://aclanthology.org/2024.osact-1.7/} {The multilingual corpus of world{'}s constitutions ({MCWC})}.
\newblock In \emph{Proceedings of the 6th Workshop on Open-Source Arabic Corpora and Processing Tools (OSACT) with Shared Tasks on Arabic LLMs Hallucination and Dialect to MSA Machine Translation @ LREC-COLING 2024}, pages 57--66, Torino, Italia. ELRA and ICCL.

\bibitem[{Elmadani et~al.(2025)Elmadani, Habash, and Taha-Thomure}]{elmadani-etal-2025-large}
Khalid~N. Elmadani, Nizar Habash, and Hanada Taha-Thomure. 2025.
\newblock \href {https://doi.org/10.18653/v1/2025.findings-acl.842} {A large and balanced corpus for fine-grained {A}rabic readability assessment}.
\newblock In \emph{Findings of the Association for Computational Linguistics: ACL 2025}, pages 16376--16400, Vienna, Austria. Association for Computational Linguistics.

\bibitem[{Elshabrawy et~al.(2023)Elshabrawy, AbuOdeh, Inoue, and Habash}]{elshabrawy-etal-2023-camelparser2}
Ahmed Elshabrawy, Muhammed AbuOdeh, Go~Inoue, and Nizar Habash. 2023.
\newblock \href {https://doi.org/10.18653/v1/2023.arabicnlp-1.15} {{C}amel{P}arser2.0: A state-of-the-art dependency parser for {A}rabic}.
\newblock In \emph{Proceedings of ArabicNLP 2023}, pages 170--180, Singapore (Hybrid). Association for Computational Linguistics.

\bibitem[{Frohmann et~al.(2024)Frohmann, Sterner, Vuli{\'c}, Minixhofer, and Schedl}]{frohmann-etal-2024-segment}
Markus Frohmann, Igor Sterner, Ivan Vuli{\'c}, Benjamin Minixhofer, and Markus Schedl. 2024.
\newblock \href {https://doi.org/10.18653/v1/2024.emnlp-main.665} {Segment any text: A universal approach for robust, efficient and adaptable sentence segmentation}.
\newblock In \emph{Proceedings of the 2024 Conference on Empirical Methods in Natural Language Processing}, pages 11908--11941, Miami, Florida, USA. Association for Computational Linguistics.

\bibitem[{Habash et~al.(2022)Habash, AbuOdeh, Taji, Faraj, El~Gizuli, and Kallas}]{habash-etal-2022-camel}
Nizar Habash, Muhammed AbuOdeh, Dima Taji, Reem Faraj, Jamila El~Gizuli, and Omar Kallas. 2022.
\newblock \href {https://aclanthology.org/2022.lrec-1.286/} {Camel treebank: An open multi-genre {A}rabic dependency treebank}.
\newblock In \emph{Proceedings of the Thirteenth Language Resources and Evaluation Conference}, pages 2672--2681, Marseille, France. European Language Resources Association.

\bibitem[{Habash and Palfreyman(2022)}]{habash-palfreyman-2022-zaebuc}
Nizar Habash and David Palfreyman. 2022.
\newblock \href {https://aclanthology.org/2022.lrec-1.9/} {{ZAEBUC}: An annotated {A}rabic-{E}nglish bilingual writer corpus}.
\newblock In \emph{Proceedings of the Thirteenth Language Resources and Evaluation Conference}, pages 79--88, Marseille, France. European Language Resources Association.

\bibitem[{Habash et~al.(2007)Habash, Soudi, and Buckwalter}]{Habash:2007:arabic-transliteration}
Nizar Habash, Abdelhadi Soudi, and Tim Buckwalter. 2007.
\newblock {On {{A}rabic} Transliteration}.
\newblock In A.~van~den Bosch and A.~Soudi, editors, \emph{{A}rabic Computational Morphology: Knowledge-based and Empirical Methods}, pages 15--22. Springer, Netherlands.

\bibitem[{Hadrich~Belguith et~al.(2005)Hadrich~Belguith, Baccour, and Ghassan}]{hadrich-belguith-etal-2005-segmentation}
Lamia Hadrich~Belguith, Leila Baccour, and Mourad Ghassan. 2005.
\newblock \href {https://aclanthology.org/2005.jeptalnrecital-court.12/} {Segmentation de textes arabes bas{\'e}e sur l{'}analyse contextuelle des signes de ponctuations et de certaines particules}.
\newblock In \emph{Actes de la 12{\`e}me conf{\'e}rence sur le Traitement Automatique des Langues Naturelles. Articles courts}, pages 451--456, Dourdan, France. ATALA.

\bibitem[{Honnibal et~al.(2020)Honnibal, Montani, Van~Landeghem, and Boyd}]{honnibal2020spacy}
Matthew Honnibal, Ines Montani, Sofie Van~Landeghem, and Adriane Boyd. 2020.
\newblock \href {https://doi.org/10.5281/zenodo.1212303} {spacy: Industrial-strength natural language processing in python}.

\bibitem[{Inoue et~al.(2021)Inoue, Alhafni, Baimukan, Bouamor, and Habash}]{inoue-etal-2021-interplay}
Go~Inoue, Bashar Alhafni, Nurpeiis Baimukan, Houda Bouamor, and Nizar Habash. 2021.
\newblock \href {https://aclanthology.org/2021.wanlp-1.10/} {The interplay of variant, size, and task type in {A}rabic pre-trained language models}.
\newblock In \emph{Proceedings of the Sixth Arabic Natural Language Processing Workshop}, pages 92--104, Kyiv, Ukraine (Virtual). Association for Computational Linguistics.

\bibitem[{Keskes et~al.(2012)Keskes, Benamara, and Belguith}]{keskes-etal-2012-clause}
Iskandar Keskes, Farah Benamara, and Lamia~Hadrich Belguith. 2012.
\newblock \href {https://aclanthology.org/L12-1559/} {Clause-based discourse segmentation of {A}rabic texts}.
\newblock In \emph{Proceedings of the Eighth International Conference on Language Resources and Evaluation ({LREC}'12)}, pages 2826--2832, Istanbul, Turkey. European Language Resources Association (ELRA).

\bibitem[{Khalil et~al.(2018)Khalil, Saddiki, Habash, and Alfalasi}]{Khalil:2018:leveled}
Muhamed~Al Khalil, Hind Saddiki, Nizar Habash, and Latifa Alfalasi. 2018.
\newblock {A Leveled Reading Corpus of Modern Standard {A}rabic}.
\newblock In \emph{Proceedings of the Language Resources and Evaluation Conference (LREC)}, Miyazaki, Japan.

\bibitem[{Kiss and Strunk(2006)}]{kissandstrunk2006}
Tibor Kiss and Jan Strunk. 2006.
\newblock \href {https://doi.org/10.1162/coli.2006.32.4.485} {Unsupervised multilingual sentence boundary detection}.
\newblock \emph{Computational Linguistics}, 32(4):485--525.

\bibitem[{Koto et~al.(2024)Koto, Li, Shatnawi, Doughman, Sadallah, Alraeesi, Almubarak, Alyafeai, Sengupta, Shehata, Habash, Nakov, and Baldwin}]{koto-etal-2024-arabicmmlu}
Fajri Koto, Haonan Li, Sara Shatnawi, Jad Doughman, Abdelrahman Sadallah, Aisha Alraeesi, Khalid Almubarak, Zaid Alyafeai, Neha Sengupta, Shady Shehata, Nizar Habash, Preslav Nakov, and Timothy Baldwin. 2024.
\newblock \href {https://doi.org/10.18653/v1/2024.findings-acl.334} {{A}rabic{MMLU}: Assessing massive multitask language understanding in {A}rabic}.
\newblock In \emph{Findings of the Association for Computational Linguistics: ACL 2024}, pages 5622--5640, Bangkok, Thailand. Association for Computational Linguistics.

\bibitem[{Lison and Tiedemann(2016)}]{lison-tiedemann-2016-opensubtitles2016}
Pierre Lison and J{\"o}rg Tiedemann. 2016.
\newblock \href {https://aclanthology.org/L16-1147/} {{O}pen{S}ubtitles2016: Extracting large parallel corpora from movie and {TV} subtitles}.
\newblock In \emph{Proceedings of the Tenth International Conference on Language Resources and Evaluation ({LREC}'16)}, pages 923--929, Portoro{\v{z}}, Slovenia. European Language Resources Association (ELRA).

\bibitem[{Liu et~al.(2026)Liu, Li, and Xu}]{liu-etal-2026-think}
Zhichen Liu, Yongyuan Li, and Yang Xu. 2026.
\newblock \href {https://doi.org/10.18653/v1/2026.acl-long.104} {Think in sentences: Explicit sentence boundaries enhance language model{'}s capabilities}.
\newblock In \emph{Proceedings of the 64th Annual Meeting of the {A}ssociation for {C}omputational {L}inguistics (Volume 1: Long Papers)}, pages 2275--2288, San Diego, California, United States. Association for Computational Linguistics.

\bibitem[{Mekki et~al.(2022)Mekki, Zribi, Ellouze, and Belguith}]{Mekki2022}
Asma Mekki, In{\`e}s Zribi, Mariem Ellouze, and Lamia~Hadrich Belguith. 2022.
\newblock \href {https://doi.org/10.1007/s10579-021-09538-4} {Sentence boundary detection of various forms of tunisian arabic}.
\newblock \emph{Language Resources and Evaluation}, 56(1):357--385.

\bibitem[{Minixhofer et~al.(2023)Minixhofer, Pfeiffer, and Vuli{\'c}}]{minixhofer-etal-2023-wheres}
Benjamin Minixhofer, Jonas Pfeiffer, and Ivan Vuli{\'c}. 2023.
\newblock \href {https://doi.org/10.18653/v1/2023.acl-long.398} {Where{'}s the point? self-supervised multilingual punctuation-agnostic sentence segmentation}.
\newblock In \emph{Proceedings of the 61st Annual Meeting of the Association for Computational Linguistics (Volume 1: Long Papers)}, pages 7215--7235, Toronto, Canada. Association for Computational Linguistics.

\bibitem[{Mohit et~al.(2014)Mohit, Rozovskaya, Habash, Zaghouani, and Obeid}]{mohit-etal-2014-first}
Behrang Mohit, Alla Rozovskaya, Nizar Habash, Wajdi Zaghouani, and Ossama Obeid. 2014.
\newblock \href {https://doi.org/10.3115/v1/W14-3605} {The first {QALB} shared task on automatic text correction for {A}rabic}.
\newblock In \emph{Proceedings of the {EMNLP} 2014 Workshop on {A}rabic Natural Language Processing ({ANLP})}, pages 39--47, Doha, Qatar. Association for Computational Linguistics.

\bibitem[{Nivre and Fang(2017)}]{nivre-fang-2017-universal}
Joakim Nivre and Chiao-Ting Fang. 2017.
\newblock \href {https://aclanthology.org/W17-0411/} {{U}niversal {D}ependency evaluation}.
\newblock In \emph{Proceedings of the {N}o{D}a{L}i{D}a 2017 Workshop on Universal Dependencies ({UDW} 2017)}, pages 86--95, Gothenburg, Sweden. Association for Computational Linguistics.

\bibitem[{Obeid et~al.(2020)Obeid, Zalmout, Khalifa, Taji, Oudah, Alhafni, Inoue, Eryani, Erdmann, and Habash}]{obeid-etal-2020-camel}
Ossama Obeid, Nasser Zalmout, Salam Khalifa, Dima Taji, Mai Oudah, Bashar Alhafni, Go~Inoue, Fadhl Eryani, Alexander Erdmann, and Nizar Habash. 2020.
\newblock \href {https://aclanthology.org/2020.lrec-1.868/} {{CAM}e{L} tools: An open source python toolkit for {A}rabic natural language processing}.
\newblock In \emph{Proceedings of the Twelfth Language Resources and Evaluation Conference}, pages 7022--7032, Marseille, France. European Language Resources Association.

\bibitem[{OpenAI et~al.(2024)OpenAI, Achiam, Adler, Agarwal, Ahmad, Akkaya, Aleman, Almeida, Altenschmidt, Altman, Anadkat, Avila, Babuschkin, Balaji, Balcom, Baltescu, Bao, Bavarian, Belgum, Bello, Berdine, Bernadett-Shapiro, Berner, Bogdonoff, Boiko, Boyd, Brakman, Brockman, Brooks, Brundage, Button, Cai, Campbell, Cann, Carey, Carlson, Carmichael, Chan, Chang, Chantzis, Chen, Chen, Chen, Chen, Chen, Chess, Cho, Chu, Chung, Cummings, Currier, Dai, Decareaux, Degry, Deutsch, Deville, Dhar, Dohan, Dowling, Dunning, Ecoffet, Eleti, Eloundou, Farhi, Fedus, Felix, Fishman, Forte, Fulford, Gao, Georges, Gibson, Goel, Gogineni, Goh, Gontijo-Lopes, Gordon, Grafstein, Gray, Greene, Gross, Gu, Guo, Hallacy, Han, Harris, He, Heaton, Heidecke, Hesse, Hickey, Hickey, Hoeschele, Houghton, Hsu, Hu, Hu, Huizinga, Jain, Jain, Jang, Jiang, Jiang, Jin, Jin, Jomoto, Jonn, Jun, Kaftan, Łukasz Kaiser, Kamali, Kanitscheider, Keskar, Khan, Kilpatrick, Kim, Kim, Kim, Kirchner, Kiros, Knight, Kokotajlo, Łukasz Kondraciuk,
  Kondrich, Konstantinidis, Kosic, Krueger, Kuo, Lampe, Lan, Lee, Leike, Leung, Levy, Li, Lim, Lin, Lin, Litwin, Lopez, Lowe, Lue, Makanju, Malfacini, Manning, Markov, Markovski, Martin, Mayer, Mayne, McGrew, McKinney, McLeavey, McMillan, McNeil, Medina, Mehta, Menick, Metz, Mishchenko, Mishkin, Monaco, Morikawa, Mossing, Mu, Murati, Murk, Mély, Nair, Nakano, Nayak, Neelakantan, Ngo, Noh, Ouyang, O'Keefe, Pachocki, Paino, Palermo, Pantuliano, Parascandolo, Parish, Parparita, Passos, Pavlov, Peng, Perelman, de~Avila Belbute~Peres, Petrov, de~Oliveira~Pinto, Michael, Pokorny, Pokrass, Pong, Powell, Power, Power, Proehl, Puri, Radford, Rae, Ramesh, Raymond, Real, Rimbach, Ross, Rotsted, Roussez, Ryder, Saltarelli, Sanders, Santurkar, Sastry, Schmidt, Schnurr, Schulman, Selsam, Sheppard, Sherbakov, Shieh, Shoker, Shyam, Sidor, Sigler, Simens, Sitkin, Slama, Sohl, Sokolowsky, Song, Staudacher, Such, Summers, Sutskever, Tang, Tezak, Thompson, Tillet, Tootoonchian, Tseng, Tuggle, Turley, Tworek, Uribe, Vallone,
  Vijayvergiya, Voss, Wainwright, Wang, Wang, Wang, Ward, Wei, Weinmann, Welihinda, Welinder, Weng, Weng, Wiethoff, Willner, Winter, Wolrich, Wong, Workman, Wu, Wu, Wu, Xiao, Xu, Yoo, Yu, Yuan, Zaremba, Zellers, Zhang, Zhang, Zhao, Zheng, Zhuang, Zhuk, and Zoph}]{openai2024gpt4technicalreport}
OpenAI, Josh Achiam, Steven Adler, Sandhini Agarwal, Lama Ahmad, Ilge Akkaya, Florencia~Leoni Aleman, Diogo Almeida, Janko Altenschmidt, Sam Altman, Shyamal Anadkat, Red Avila, Igor Babuschkin, Suchir Balaji, Valerie Balcom, Paul Baltescu, Haiming Bao, Mohammad Bavarian, Jeff Belgum, and 262 others. 2024.
\newblock \href {https://arxiv.org/abs/2303.08774} {Gpt-4 technical report}.
\newblock \emph{Preprint}, arXiv:2303.08774.

\bibitem[{Pevzner and Hearst(2002)}]{pevzner-hearst-2002-critique}
Lev Pevzner and Marti~A. Hearst. 2002.
\newblock \href {https://doi.org/10.1162/089120102317341756} {A critique and improvement of an evaluation metric for text segmentation}.
\newblock \emph{Computational Linguistics}, 28(1):19--36.

\bibitem[{Qiu et~al.(2025)Qiu, Li, Zhang, Zhang, Zhang, Zhang, and Yu}]{qiu2025sentencelevelrewardmodelgeneralize}
Wenjie Qiu, Yi-Chen Li, Xuqin Zhang, Tianyi Zhang, Yihang Zhang, Zongzhang Zhang, and Yang Yu. 2025.
\newblock \href {https://arxiv.org/abs/2503.04793} {Sentence-level reward model can generalize better for aligning llm from human preference}.
\newblock \emph{Preprint}, arXiv:2503.04793.

\bibitem[{{Qwen Team}(2026)}]{qwen3.6-27b}
{Qwen Team}. 2026.
\newblock \href {https://qwen.ai/blog?id=qwen3.6-27b} {{Qwen3.6-27B}: Flagship-level coding in a {27B} dense model}.

\bibitem[{Retkowski and Waibel(2026)}]{retkowski-etal-2026-paragraph}
Fabian Retkowski and Alexander Waibel. 2026.
\newblock \href {https://doi.org/10.63317/3eczsids4mek} {Paragraph segmentation revisited: Towards a standard task for structuring speech}.
\newblock In \emph{Proceedings of the Fifteenth Language Resources and Evaluation Conference (LREC 2026)}, pages 747--759, Palma, Mallorca, Spain. European Language Resources Association (ELRA).

\bibitem[{Sadvilkar and Neumann(2020)}]{sadvilkar-neumann-2020-pysbd}
Nipun Sadvilkar and Mark Neumann. 2020.
\newblock \href {https://doi.org/10.18653/v1/2020.nlposs-1.15} {{P}y{SBD}: Pragmatic sentence boundary disambiguation}.
\newblock In \emph{Proceedings of Second Workshop for NLP Open Source Software (NLP-OSS)}, pages 110--114, Online. Association for Computational Linguistics.

\bibitem[{Singh et~al.(2026)Singh, Fry, Perelman, Tart, Ganesh, El-Kishky, McLaughlin, Low, Ostrow, Ananthram, Nathan, Luo, Helyar, Madry, Efremov, Spyra, Baker-Whitcomb, Beutel, Karpenko, Makelov, Neitz, Wei, Barr, Kirchmeyer, Ivanov, Christakis, Gillespie, Tam, Bennett, Wan, Huang, Sandjideh, Yang, Kumar, Saraiva, Vallone, Gheorghe, Garcia, Braunstein, Liu, Schmidt, Mereskin, Mishchenko, Applebaum, Rogerson, Rajan, Wei, Kotha, Srivastava, Agrawal, Vijayvergiya, Tyra, Nair, Nayak, Eggers, Ji, Hoover, Chen, Chen, Barak, Minaiev, Hao, Baker, Lightcap, McKinzie, Wang, Quinn, Fioca, Hsu, Yang, Yu, Zhang, Brenner, Zetino, Raymond, Lugaresi, Paz, Hudson, Whitney, Li, Chen, Cole, Voss, Ding, Shen, Huang, Colby, Hallacy, Koch, Lu, Kaplan, Kim, Minott-Henriques, Frey, Yu, Czarnecki, Reid, Wei, Decareaux, Scheau, Zhang, Forbes, Tang, Goldberg, Roberts, Palmie, Kappler, Levine, Wright, Leo, Lin, Robinson, Grabb, Chen, Lim, Salama, Bhattacharjee, Tsipras, Li, Yu, Strouse, Williams, Hunn, Bayes, Arbus, Akyurek, Le,
  Widmann, Yani, Proehl, Sert, Cheung, Schwartz, Han, Jiang, Mitchell, Sigler, Wallace, Ritter, Kavanaugh, Mays, Nikishin, Li, Such, de~Avila Belbute~Peres, Raso, Bekerman, Tsimpourlas, Chantzis, Song, Zhang, Raila, McGrath, Briggs, Yang, Parascandolo, Chabot, Kim, Zhao, Valiant, Leclerc, Salman, Wang, Sheng, Jiang, Wang, Jin, Sikchi, Schmidt, Aspegren, Chen, Qiu, Lightman, Covert, Kivlichan, Silber, Sohl, Hammoud, Clavera, Lan, Akkaya, Kostrikov, Kofman, Etinger, Singal, Hehir, Huh, Pan, Wilczynski, Pachocki, Lee, Quinn, Kiros, Kalra, Samaroo, Wang, Wolfe, Chen, Wang, Harb, Han, Wang, Zhao, Chen, Yang, Tworek, Chand, Landon, Liang, Lin, Liu, Wang, Tang, Yin, Jang, Morris, Flynn, Ferstad, Heidecke, Fishbein, Hallman, Grant, Chien, Gordon, Park, Liss, Kraaijeveld, Guay, Mo, Lawson, McGrath, Vendrow, Jiao, Lee, Steele, Wang, Mao, Chen, Hayashi, Xiao, Salahi, Wu, Sekhri, Sharma, Singhal, Li, Nguyen, Gu-Lemberg, King, Liu, Stone, Yu, Ying, Georgiev, Lim, Tirumala, Miller, Ahmad, Lv, Clare, Fauconnet, Itow, Yang,
  Romaniuk, Anise, Byron, Pathak, Maksin, Lo, Ho, Jing, Wu, Xiong, Mamitsuka, Yang, McCallum, Held, Bourgeois, Engstrom, Kuhn, Feuvrier, Zhang, Switzer, Kondraciuk, Kaiser, Joglekar, Singh, Shah, Stratta, Williams, Chen, Sun, Cayton, Li, Zhang, Aljubeh, Nichols, Haines, Schwarzer, Gupta, Shah, Guan, Huang, Dong, Wang, Glaese, Carroll, Lampe, Malek, Sharman, Zhang, Wang, Pokrass, Florian, Pavlov, Wang, Chen, Wang, Feng, Bavarian, Lin, Abdool, Rohaninejad, Soto, Staudacher, LaFontaine, Marwell, Liu, Preston, Turley, Ansman, Blades, Pancha, Mikhaylin, Felix, Handa, Rai, Keskar, Brown, Nachum, Boiko, Murk, Watkins, Gleeson, Mishkin, Lesiewicz, Baltescu, Belov, Zhokhov, Pronin, Guo, Thacker, Liu, Yuan, Liu, Dias, Puckett, Arora, Mullapudi, Gaon, Miyara, Song, Aggarwal, Marsan, Yemiru, Xiong, Kshirsagar, Nuttall, Tsiupa, Eldan, Wang, James, Ziv, Shu, Nigmatullin, Jain, Talaie, Altman, Arnesen, Toizer, Toyer, Miserendino, Agarwal, Yoo, Heon, Ethersmith, Grove, Taylor, Bubeck, Banesiu, Amdo, Zhao, Wu, Santurkar,
  Zhao, Chaudhuri, Krishnaswamy, Shuaiqi, Xia, Cheng, Anadkat, Fishman, Tobin, Fu, Jain, Mei, Egoian, Kim, Golden, Mah, Lin, Imm, Sharpe, Yadlowsky, Choudhry, Eum, Sanjeev, Khan, Stramer, Wang, Xin, Gogineni, Christianson, Sanders, Patwardhan, Degry, Shadwell, Fu, Gao, Garipov, Sriskandarajah, Sherbakov, Korbak, Kaftan, Hiratsuka, Wang, Song, Zhao, Peterson, Kharitonov, Chernova, Kosaraju, Kuo, Pong, Verma, Petrov, Jiang, Zhang, Zhou, Xie, Zhan, McCabe, DePue, Ellsworth, Bain, Thompson, Chen, Qi, Xiang, Shi, Dubois, Yu, Khakbaz, Wu, Qian, Lee, Chen, Zhang, Xiong, Tian, Cha, Bai, Yang, Yuan, Li, Zhang, Yang, Jin, Jiang, Wang, Wang, Liu, Stubenvoll, Dou, Wu, and Wang}]{gpt5}
Aaditya Singh, Adam Fry, Adam Perelman, Adam Tart, Adi Ganesh, Ahmed El-Kishky, Aidan McLaughlin, Aiden Low, AJ~Ostrow, Akhila Ananthram, Akshay Nathan, Alan Luo, Alec Helyar, Aleksander Madry, Aleksandr Efremov, Aleksandra Spyra, Alex Baker-Whitcomb, Alex Beutel, Alex Karpenko, and 467 others. 2026.
\newblock \href {https://arxiv.org/abs/2601.03267} {Openai gpt-5 system card}.
\newblock \emph{Preprint}, arXiv:2601.03267.

\bibitem[{Smith and Van~Dyck(1860)}]{arabicNewTestament}
Eli Smith and Cornelius Van~Dyck. 1860.
\newblock \emph{{New Testament (Arabic Translation)}}.

\bibitem[{Smith and Van~Dyck(1865)}]{arabicOldTestament}
Eli Smith and Cornelius Van~Dyck. 1865.
\newblock \emph{{Old Testament (Arabic Translation)}}.

\bibitem[{Taha-Thomure(2007)}]{poetry-and-news}
Hanada Taha-Thomure. 2007.
\newblock \emph{Poems and News ({\footnotesize \<أشعار وأخبار>})}.
\newblock Educational Book House ({\footnotesize \<دار الكتاب التربوي للنشر والتوزيع>)}.

\bibitem[{Takezawa et~al.(2007)Takezawa, Kikui, Mizushima, and Sumita}]{takezawa-etal-2007-multilingual}
Toshiyuki Takezawa, Genichiro Kikui, Masahide Mizushima, and Eiichiro Sumita. 2007.
\newblock \href {https://aclanthology.org/O07-5005/} {Multilingual spoken language corpus development for communication research}.
\newblock In \emph{International Journal of Computational Linguistics {\&} {C}hinese Language Processing, Volume 12, Number 3, September 2007: Special Issue on Invited Papers from {ISCSLP} 2006}, pages 303--324.

\bibitem[{TEAM et~al.(2026)TEAM, Abbas, Ahmad, Ahmad, Al-Homaid, Al-Nuaimi, Altinisik, Asgari, Chawla, Chowdhury, Dalvi, Darwish, Durrani, Elfeky, Elmagarmid, Eltabakh, Ersoy, Fatehkia, Hashim, Hawasly, Hefeeda, Husaini, Isufaj, Jung, Lachemat, Lucas, Mohamed, Mohiuddin, Mousi, Mubarak, Musleh, Ouzzani, Sadeghi, Sencar, Shinoy, Sinan, and Zhang}]{fanarteam2026fanar20arabicgenerative}
FANAR TEAM, Ummar Abbas, Mohammad~Shahmeer Ahmad, Minhaj Ahmad, Abdulaziz Al-Homaid, Anas Al-Nuaimi, Enes Altinisik, Ehsaneddin Asgari, Sanjay Chawla, Shammur Chowdhury, Fahim Dalvi, Kareem Darwish, Nadir Durrani, Mohamed Elfeky, Ahmed Elmagarmid, Mohamed Eltabakh, Asim Ersoy, Masoomali Fatehkia, Mohammed~Qusay Hashim, and 18 others. 2026.
\newblock \href {https://arxiv.org/abs/2603.16397} {Fanar 2.0: Arabic generative ai stack}.
\newblock \emph{Preprint}, arXiv:2603.16397.

\bibitem[{Team(2024)}]{qwen2.5}
Qwen Team. 2024.
\newblock \href {https://qwenlm.github.io/blog/qwen2.5/} {Qwen2.5: A party of foundation models}.

\bibitem[{Tufail(1150)}]{tufail:hayy}
Ibn Tufail. 1150.
\newblock \href {https://www.hindawi.org/books/90463596/} {\emph{{Hayy ibn Yaqdhan}}}.
\newblock Hindawi.

\bibitem[{Unknown(12th century)}]{ArabianNights}
Unknown. 12th century.
\newblock \emph{One Thousand and One Nights}.

\bibitem[{Wicks and Post(2021)}]{wicks-post-2021-unified}
Rachel Wicks and Matt Post. 2021.
\newblock \href {https://doi.org/10.18653/v1/2021.acl-long.309} {A unified approach to sentence segmentation of punctuated text in many languages}.
\newblock In \emph{Proceedings of the 59th Annual Meeting of the Association for Computational Linguistics and the 11th International Joint Conference on Natural Language Processing (Volume 1: Long Papers)}, pages 3995--4007, Online. Association for Computational Linguistics.

\bibitem[{Xu and Ma(2025)}]{xu-ma-2025-llm}
Nan Xu and Xuezhe Ma. 2025.
\newblock \href {https://doi.org/10.18653/v1/2025.naacl-long.172} {{LLM} the genius paradox: A linguistic and math expert{'}s struggle with simple word-based counting problems}.
\newblock In \emph{Proceedings of the 2025 Conference of the Nations of the Americas Chapter of the Association for Computational Linguistics: Human Language Technologies (Volume 1: Long Papers)}, pages 3344--3370, Albuquerque, New Mexico. Association for Computational Linguistics.

\bibitem[{Yagi et~al.(2024)Yagi, Fareh, Elnagar, Balajeed, El-mneizel, and Al-Badawi}]{yagi2024}
Sane Yagi, Shehdeh Fareh, Ashraf Elnagar, Mariam Balajeed, Abdalla El-mneizel, and Mohammad Al-Badawi. 2024.
\newblock \href {https://doi.org/10.1080/23311983.2024.2303818} {Is {A}rabic punctuation rule-governed?}
\newblock \emph{Cogent Arts \& Humanities}, 11(1):2303818.

\bibitem[{Yang et~al.(2024)Yang, Yang, Hui, Zheng, Yu, Zhou, Li, Li, Liu, Huang, Dong, Wei, Lin, Tang, Wang, Yang, Tu, Zhang, Ma, Xu, Zhou, Bai, He, Lin, Dang, Lu, Chen, Yang, Li, Xue, Ni, Zhang, Wang, Peng, Men, Gao, Lin, Wang, Bai, Tan, Zhu, Li, Liu, Ge, Deng, Zhou, Ren, Zhang, Wei, Ren, Fan, Yao, Zhang, Wan, Chu, Liu, Cui, Zhang, and Fan}]{qwen2}
An~Yang, Baosong Yang, Binyuan Hui, Bo~Zheng, Bowen Yu, Chang Zhou, Chengpeng Li, Chengyuan Li, Dayiheng Liu, Fei Huang, Guanting Dong, Haoran Wei, Huan Lin, Jialong Tang, Jialin Wang, Jian Yang, Jianhong Tu, Jianwei Zhang, Jianxin Ma, and 40 others. 2024.
\newblock Qwen2 technical report.
\newblock \emph{arXiv preprint arXiv:2407.10671}.

\bibitem[{Zaghouani and Awad(2016)}]{zaghouani2016toward}
Wajdi Zaghouani and Dana Awad. 2016.
\newblock Toward an {A}rabic punctuated corpus: Annotation guidelines and evaluation.
\newblock In \emph{The 2nd Workshop on Arabic Corpora and Processing Tools 2016 Theme: Social Media}, page~22.

\bibitem[{Zaghouani et~al.(2014)Zaghouani, Mohit, Habash, Obeid, Tomeh, Rozovskaya, Farra, Alkuhlani, and Oflazer}]{zaghouani-etal-2014-large}
Wajdi Zaghouani, Behrang Mohit, Nizar Habash, Ossama Obeid, Nadi Tomeh, Alla Rozovskaya, Noura Farra, Sarah Alkuhlani, and Kemal Oflazer. 2014.
\newblock \href {https://aclanthology.org/L14-1721/} {Large scale {A}rabic error annotation: Guidelines and framework}.
\newblock In \emph{Proceedings of the Ninth International Conference on Language Resources and Evaluation ({LREC}'14)}, Reykjavik, Iceland. European Language Resources Association (ELRA).

\bibitem[{Zheng et~al.(2025)Zheng, Liu, Li, Chen, Yu, Gao, Dang, Liu, Men, Yang, Zhou, and Lin}]{zheng2025groupsequencepolicyoptimization}
Chujie Zheng, Shixuan Liu, Mingze Li, Xiong-Hui Chen, Bowen Yu, Chang Gao, Kai Dang, Yuqiong Liu, Rui Men, An~Yang, Jingren Zhou, and Junyang Lin. 2025.
\newblock \href {https://arxiv.org/abs/2507.18071} {Group sequence policy optimization}.
\newblock \emph{Preprint}, arXiv:2507.18071.

\end{thebibliography}
\clearpage
\appendix
\section{Supplemental Statistics}
\subsection{Dataset Splits}
\label{sec:dataset_splits}
\begin{table}[t]
\centering
\tabcolsep3pt
\begin{tabular}{lcccc}
\toprule
\textbf{Split} & \textbf{\#Docs} &
\textbf{\#Paras} &
\textbf{\#Sents} & \textbf{\#Tokens} \\
\midrule
Train & 1,703 & 38,159 & 96,129 & 1.1M \\
Dev   & 222  & 5,066  & 12,985  & 159K \\
Test  & 262  & 5,025  & 12,509  & 154K \\
\midrule\midrule
Total  & 2,187 & 48,250 & 121,623 & 1.5M \\
\bottomrule
\end{tabular}
\caption{{\AraSEG} split statistics by documents (Docs), paragraphs (Paras), sentences (Sents), and tokens.}
\label{tab:corpus-splits}
\end{table}

Table~\ref{tab:corpus-splits} reports corpus statistics for each split. We preserve the original train/dev/test splits of CAMeLTB and BAREC. In both datasets, some Quran chapters are divided across multiple documents due to their length; we retain these document boundaries and do not split any newly added chapters.

\subsection{Source-level Dataset Statistics}
\label{sec:appendix_dataset_stats}
\begin{table*}[!t]
\centering
\small
\setlength{\tabcolsep}{3.5pt}
\begin{tabular}{llccccccccc}
\toprule
\textbf{Genre} & \textbf{Source} & \textbf{\#Docs} & \textbf{\#Paras} & \textbf{\#Sents} & \textbf{\#Tokens} & \textbf{\#Words} & \textbf{\#Pnx} & \textbf{\#PnxClus} & \textbf{PnxDen} & \textbf{Sent. Len.} \\
\midrule

\multirow{6}{*}{Children}
 & Green Library & 58 & 1.1K & 2.8K & 42.7K & 35.4K & 10.1K & 7.3K & 23.6\% & 15.0 \\
 & Kashkul & 15 & 376 & 404 & 2.5K & 2.2K & 307 & 269 & 12.2\% & 6.2 \\
 & Majed & 294 & 6.8K & 11.9K & 119.5K & 100.6K & 23.7K & 19K & 19.9\% & 10.1 \\
 & Mama Bread & 1 & 23 & 38 & 437 & 381 & 87 & 56 & 19.9\% & 11.5 \\
 & Spacetoon & 51 & 1.1K & 1.1K & 5K & 4.2K & 962 & 778 & 19.4\% & 4.7 \\
 & chatGPT & 3 & 449 & 449 & 2.5K & 2.1K & 449 & 449 & 17.6\% & 5.7 \\
\cmidrule(lr){2-11}
 & {Total} & {422} & {9.8K} & {16.7K} & {172.7K} & {144.8K} & {35.6K} & {27.8K} & {20.6\%} & {10.4} \\

\midrule

\multirow{5}{*}{Education}
 & ALC & 16 & 86 & 727 & 9.2K & 8.2K & 1.1K & 1K & 11.7\% & 12.6 \\
 & ArabicMMLU & 344 & 6.8K & 31.2K & 203.3K & 144.1K & 66.9K & 59.2K & 32.9\% & 6.5 \\
 & BTEC & 20 & 2K & 2K & 15.9K & 13.1K & 2.9K & 2.8K & 18.0\% & 8.0 \\
 & Curriculum & 125 & 9.7K & 15.3K & 129.1K & 108.6K & 22.4K & 20.4K & 17.4\% & 8.4 \\
 & ZAEBUC & 100 & 166 & 1.1K & 15.8K & 14.5K & 1.3K & 1.3K & 8.1\% & 14.2 \\
\cmidrule(lr){2-11}
 & {Total} & {605} & {18.8K} & {50.3K} & {373.3K} & {288.5K} & {94.6K} & {84.7K} & {25.3\%} & {7.4} \\

\midrule

\multirow{5}{*}{Literature}
 & Arabian Nights & 35 & 44 & 1.1K & 11.8K & 11.4K & 407 & 407 & 3.4\% & 10.3 \\
 & Hayy & 20 & 379 & 1.2K & 19.7K & 18.3K & 1.4K & 1.4K & 7.0\% & 16.4 \\
 & Hindawi & 269 & 5K & 13.5K & 228.2K & 198.5K & 35.3K & 29.8K & 15.5\% & 16.9 \\
 & Kalima & 62 & 1K & 2.9K & 44.9K & 38.9K & 6.4K & 5.9K & 14.3\% & 15.4 \\
 & Sara & 20 & 1.5K & 2.2K & 34.8K & 30.9K & 4.5K & 4K & 12.9\% & 16.1 \\
\cmidrule(lr){2-11}
 & {Total} & {406} & {8K} & {21K} & {339.4K} & {298K} & {48.1K} & {41.5K} & {14.2\%} & {16.2} \\

\midrule

\multirow{3}{*}{Media}
 & QALB & 20 & 217 & 923 & 11.2K & 9.7K & 1.7K & 1.5K & 15.5\% & 12.2 \\
 & Subtitles & 11 & 567 & 567 & 3.5K & 3K & 588 & 495 & 16.8\% & 6.2 \\
 & WikiNews & 70 & 392 & 996 & 18.2K & 16.6K & 1.8K & 1.6K & 9.7\% & 18.2 \\
\cmidrule(lr){2-11}
 & {Total} & {101} & {1.2K} & {2.5K} & {32.9K} & {29.3K} & {4.1K} & {3.6K} & {12.4\%} & {13.2} \\

\midrule

\multirow{4}{*}{Poetry}
 & Hanada & 1 & 388 & 388 & 1.3K & 1.2K & 119 & 103 & 9.5\% & 3.2 \\
 & Hanging Odes & 10 & 784 & 784 & 7.4K & 7.3K & 51 & 50 & 0.7\% & 9.4 \\
 & Lang. Sings & 18 & 424 & 424 & 2.8K & 2.2K & 702 & 599 & 25.5\% & 6.5 \\
 & Al Issa & 1 & 100 & 100 & 333 & 316 & 32 & 17 & 9.6\% & 3.3 \\
\cmidrule(lr){2-11}
 & {Total} & {30} & {1.7K} & {1.7K} & {11.7K} & {10.9K} & {904} & {769} & {7.7\%} & {6.9} \\

\midrule

\multirow{2}{*}{Politics}
 & Constitutions & 16 & 5.8K & 9.6K & 145.5K & 130.5K & 15.3K & 15K & 10.5\% & 15.1 \\
 & UN DEC HR & 1 & 79 & 91 & 1.4K & 1.2K & 227 & 227 & 15.9\% & 15.7 \\
\cmidrule(lr){2-11}
 & {Total} & {17} & {5.9K} & {9.7K} & {147K} & {131.7K} & {15.5K} & {15.2K} & {10.6\%} & {15.1} \\

\midrule

\multirow{4}{*}{Religion}
 & Hadith & 134 & 135 & 1.2K & 12.3K & 10.5K & 1.9K & 1.8K & 15.3\% & 10.3 \\
 & NT & 65 & 507 & 3.2K & 53.3K & 44.8K & 10.7K & 8.5K & 20.0\% & 16.5 \\
 & OT & 88 & 577 & 3.1K & 54.3K & 46.1K & 9.8K & 8.1K & 18.1\% & 17.6 \\
 & Quran & 151 & 151 & 6.2K & 96.1K & 83.7K & 12.5K & 12.5K & 13.0\% & 15.4 \\
\cmidrule(lr){2-11}
 & {Total} & {438} & {1.4K} & {13.7K} & {216K} & {185.1K} & {34.8K} & {30.9K} & {16.1\%} & {15.7} \\

\midrule

\multirow{1}{*}{Wiki}
 & Wikipedia & 168 & 1.6K & 6.1K & 128.6K & 113.1K & 17.7K & 15.5K & 13.8\% & 21.2 \\
\cmidrule(lr){2-11}
 & {Total} & {168} & {1.6K} & {6.1K} & {128.6K} & {113.1K} & {17.7K} & {15.5K} & {13.8\%} & {21.2} \\

\midrule\midrule

\multirow{1}{*}{\textbf{Corpus Total}}
 & & \textbf{2.2K} & \textbf{48.3K} & \textbf{121.6K} & \textbf{1.4M} & \textbf{1.2M} & \textbf{251.3K} & \textbf{220.1K} & \textbf{17.7\%} & \textbf{11.7} \\

\bottomrule
\end{tabular}
\caption{{\AraSEG} statistics across genres and sources, including documents (Docs), paragraphs (Paras), sentences (Sents), total tokens (Tokens), word tokens (Words), punctuation tokens (Pnx), punctuation clusters (PnxClus), punctuation density (PnxDen), and average sentence length (Sent. Len).}
\label{tab:corpus-genre}
\end{table*}
We report the number of documents, paragraphs, sentences, tokens, word tokens, punctuation tokens, number of punctuation clusters (PnxClus), Punctuation density (PnxDen) and average sentence length (Sent. Len.) across all sources in Table~\ref{tab:corpus-genre}.

\begin{table}[t]
\centering
\setlength{\tabcolsep}{4pt}
\begin{tabular}{lccc}
\toprule
\textbf{PnxClus} & \textbf{Count} & \textbf{SegPrec} & \textbf{Type} \\
\midrule
.                & 41,277 & 82.9 & Single \\
\<،>             & 56,315 & 18.6 & Single \\
)                & 33,251 & 20.5 & Single \\
:                & 11,182 & 45.2 & Single \\
"                & 7,083  & 3.4  & Single \\
\<؟>             & 5,396  & 89.3 & Single \\
..               & 3,643  & 70.0 & Repeated \\
\<؛>             & 2,987  & 41.9 & Single \\
".               & 2,114  & 52.9 & Mixed \\
!                & 2,016  & 83.7 & Single \\
...              & 733    & 81.6 & Repeated \\
).               & 673    & 93.0 & Mixed \\
."               & 506    & 95.8 & Mixed \\
\<؟>!            & 271    & 99.3 & Mixed \\
\bottomrule
\end{tabular}
\caption{Most frequent punctuation clusters (PnxClus) in \AraSEG, along with their segmentation precision (SegPrec).}
\label{tab:pnx-precision}
\end{table}

\subsection{Overall Punctuation}
\label{sec:app_overall_pnx}
Table~\ref{tab:pnx-precision} shows counts and segmentation precision for the most frequent punctuation clusters.



\section{Dataset}
\label{sec:appendix_dataset}


\label{sec:appendix_dataset_sources}

 
We present the corpus sources in groups of their general intended purpose.

\subsection{Education}

\paragraph{Emarati Curriculum} The first five units of the UAE curriculum textbooks for the 12 grades in three subjects: Arabic language, social studies, Islamic studies \cite{Khalil:2018:leveled}.

\paragraph{ArabicMMLU} Question and answer pairs from the ArabicMMLU benchmark dataset \cite{koto-etal-2024-arabicmmlu}.

\paragraph{Zayed Arabic-English Bilingual Undergraduate Corpus (ZAEBUC)}
100 student-written articles from the Zayed University Arabic-English Bilingual Undergraduate Corpus \cite{habash-palfreyman-2022-zaebuc,alhafni-etal-2026-bilingual}.

\paragraph{Arabic Learner Corpus (ALC)}
16 L2 articles from the Arabic Learner Corpus \citep{phdthesis}.

\paragraph{Basic Travel Expressions Corpus (BTEC)}
20 documents from the MSA translation of the Basic Traveling Expression Corpus \cite{eck-hori-2005-overview,takezawa-etal-2007-multilingual,bouamor-etal-2018-madar}.

\subsection{Children}
\paragraph{ChatGPT} To add more children's materials, we ask ChatGPT to generate 200 sentences ranging from 2 to 4 words per sentence, 150 sentences ranging from 5 to 7 words per sentence and 100 sentences ranging from 8 to 10 words per sentence.\footnote{\url{https://chatgpt.com/}} Not all sentences generated by ChatGPT were correct. We discarded some sentences that were flagged by the annotators.

\paragraph{Collection of Children poems} Example of the included poems: My language sings (\<لغتي تغني>), and Poetry and news (\<أشعار وأخبار>) \cite{kashkol, poetry-and-news}.

\paragraph{Spacetoon Songs}
The opening songs of 53 children's animated series from Spacetoon channel.

\paragraph{Majed}
10 manually typed editions of Majed magazine for children from 1983 to 2019.\footnote{\url{https://archive.org/details/majid_magazine}}

\paragraph{Green Library}
58 manually typed books from the Green Library.\footnote{\url{https://archive.org/details/201409_201409}}

\subsection{Literature}

\paragraph{Hindawi} A subset of 264 books extracted from the Hindawi Foundation website across different genres.\footnote{\url{https://www.hindawi.org/books/categories/}}

\paragraph{Kalima}
The first 500 words of 62 books from Kalima project.\footnote{\url{https://alc.ae/publications/kalima/}}

\paragraph{Arabian Nights} The openings and endings of the opening narrative and the first eight nights from the Arabian Nights \cite{ArabianNights}. We extracted the text from an online forum.\footnote{\url{http://al-nada.eb2a.com/1000lela\&lela/}}

\paragraph{Hayy ibn Yaqdhan}  A subset of the philosophical novel and allegorical tale written by Ibn Tufail \cite{tufail:hayy}.
We extracted the text from the Hindawi Foundation website.\footnote{\url{https://www.hindawi.org/books/90463596/}}

\paragraph{Sara} The full text of {\it Sara}, a novel by Al-Akkad first published in 1938 \cite{akkad:sarah}. We extracted the text from the Hindawi Foundation website.\footnote{\url{https://www.hindawi.org/books/72707304/}}

\subsection{Poetry}
\paragraph{The Suspended Odes (Odes)}  The ten most celebrated poems from Pre-Islamic Arabia (\<المعلقات> Mu’allaqat).
All texts were extracted from Wikipedia.\footnote{\url{https://ar.wikipedia.org/wiki/}\<المعلقات>}

\subsection{Media}

\paragraph{Subtitles} A subset of the Arabic side of the OpenSubtitles 
dataset \cite{lison-tiedemann-2016-opensubtitles2016}.

\paragraph{QALB} 200 online comments from the Qatar Arabic Language Bank (QALB)~\cite{mohit-etal-2014-first}.

\paragraph{WikiNews} 62 Arabic articles covering politics, economics,
health, science and technology, sports, arts, and culture \cite{abdelali-etal-2016-farasa}.

\subsection{Wiki} 
\paragraph{Wikipedia} A subset of 168 Arabic Wikipedia articles covering Culture, Figures, Geography, History, Mathematics, Sciences, Society, Philosophy, Religions and Technologies.\footnote{\url{https://ar.wikipedia.org/}}

\subsection{Politics}
\paragraph{Constitutions}
The Arabic constitutions from 16 Arabic-speaking countries, collected from MCWC dataset \cite{el-haj-ezzini-2024-multilingual}.

\paragraph{UN} The Arabic translation of the Universal Declaration of Human Rights.\footnote{\url{https://www.un.org/ar/about-us/universal-declaration-of-human-rights}}

\subsection{Religion}

\paragraph{Old Testament} The Books of Gensis and Exodus~\cite{arabicOldTestament}.\footnote{\url{https://www.arabicbible.com/}\label{biblefoot}}

\paragraph{New Testament} The Books of Matthew, Mark, and Luke~\cite{arabicNewTestament}.$^{\ref{biblefoot}}$ 

\paragraph{Quran} The entire Quran~\cite{Dukes:2013:supervised}.\footnote{\url{https://corpus.quran.com/}}

\paragraph{Hadith} The first 75 Hadiths from Sahih Bukhari \cite{bukhari}.  We selected the text from the LK Hadith Corpus~\cite{Altammami:2019:Arabic}.\footnote{\url{https://github.com/ShathaTm/LK-Hadith-Corpus}}

\section{Models and Hyperparameters}
\label{sec:appendix_training}
\subsection{Compute}
Most experiments were run  on 2 A100-40GB SXM GPUs. Training runs vary by task. NP task requires around 20 mins of training time, whereas PA tasks require 40 mins of training time for CAMeLBERT models. For Parsers, we train using 1 A100-40GB SXM GPU, which runs for approximately 25 mins. For Jais-2-70B-Chat, we rent a B200 GPU, and run inference for 2 hours per setting per split using bfloat16.
\subsection{CAMeLBERT}
We use pretrained CAMeLBERT-MSA for finetuning. We train with an effective batch size of 32 for 5 epochs with learning rate of $3e^{-5}$. We use chunk size 512, and stride length 64, averaging overlapping logits. 
\subsection{Dependency Parsers}
\label{sec:dep-parsers}
We train biaffine-attention dependency parsers on silver labels generated using CamelParser 2.0. 

\paragraph{Training Data} We generate the trees that are used to explicitly model sentence segmentation by re-indexing the gold or silver trees such that all sentences belonging to the same document point to the same artificial root. 

\paragraph{Predicting Sentence Boundaries} At inference time, we run BFS on all nodes attached to the artificial root to get their subtrees, and we add a newline character after the token with the largest index in the subtree.

\subsection{Baselines}
We use default settings for all baselines.
\section{LLMs}
\subsection{Compute Costs}
For closed-source LLMs, we use \$800 of credits. For Fanar-2-27B, the API is free to use under a rate limit. 
\label{sec:appendix_llms}
\subsection{Prompts}
\label{sec:llm_prompts}
Prompt.~\ref{prompt:commercial-np} shows the prompt used for NP and NoPnx-NP tasks. Prompt.~\ref{prompt:commercial-pa} shows the prompt used for commercial LLMs for both PA and NoPnx-PA tasks. Open-source LLMs are much smaller in size, and struggle to follow the expected JSON format. Thus, we used a simpler format for them. The prompt is shown in Prompt.~\ref{prompt:simple-prompt}.
We align outputs before running evaluation to handle hallucinations.

\refstepcounter{promptcounter}
\begin{tcolorbox}[colback=white!5!white,
colframe=orange!75!black,
fonttitle=\bfseries\normalsize,
fontupper=\small,
title=Prompt.~\thepromptcounter\ Commercial LLMs NP \& NoPnx-NP System prompt
]
\label{prompt:commercial-np}

\# Task \\
You are a helpful assistant that segments Arabic Documents into sentences. Use the following guidelines to segment the document:\\
- The document is given in the "text" field.\\
- You must return a list of sentences in the "sentences" field.\\
- Do not modify the text or the sentence boundaries.\\
- Do not remove any words or punctuation marks.\\
- Do NOT remove "-" from the text.\\
- DO NOT ADD YOUR OWN PUNCTUATION MARKS.\\
- The words MUST be in the same order as they appear in the text.\\
- Do not merge punctuation marks into words, they are separate by default.\\
- You MUST return a valid JSON object in the following format:

\begin{verbatim}
# Output Format
{{
    "doc_name": "doc_name",
    "sentences":
    [
        "sentence0",
        "sentence1",
        ...
    ]
}}

# Task
Input:
{payload}
Output:
\end{verbatim}

\end{tcolorbox}

\refstepcounter{promptcounter}
\begin{tcolorbox}[colback=white!5!white,
colframe=orange!75!black,
fonttitle=\bfseries\normalsize,
fontupper=\small,
breakable,
title=Prompt.~\thepromptcounter\ Commercial LLMs PA \& NoPnx-PA System prompt
]
\label{prompt:commercial-pa}

\# Task \\
You segment Arabic text into sentences \textbf{within each paragraph only}. \\
- The input is one JSON object with \texttt{doc\_name} and \texttt{paragraphs}: a list of objects
  each with \texttt{paragraph\_id} (integer) and \texttt{text} (that paragraph only: sentences
  joined with single spaces, no newlines).\\
- For \textbf{each} paragraph independently, return \texttt{sentences}: the list of sentence
  strings for that paragraph. Do not merge or split across paragraphs.\\
- Preserve every word and token order; do not remove "-" or punctuation; do not add punctuation.\\
- You MUST return a valid JSON object in the output format below with the same
  \texttt{paragraph\_id} values as in the input.

\begin{verbatim}
# Output Format
{{
    "doc_name": "doc_name",
    "paragraphs": [
        {
            "paragraph_id": 1,
            "sentences": [
                "...",
                "..."
            ]
        },
        {
            "paragraph_id": 2,
            "sentences": [
                "..."
            ]
        }
    ]
}}

# Task
Input:
{payload}
Output:
\end{verbatim}

\end{tcolorbox}

\refstepcounter{promptcounter}
\begin{tcolorbox}[colback=white!5!white,
colframe=orange!75!black,
fonttitle=\bfseries\normalsize,
fontupper=\small,
breakable,
title=Prompt.~\thepromptcounter\ Simple System prompt
]
\label{prompt:simple-prompt}

You are an expert in Arabic text segmentation.
\textbf{INSERT} newlines between sentences in the text. Do not modify the text or make any changes to it. Only insert \textbf{NEWLINES}. DO NOT COMBINE PUNCTUATION MARKS WITH WORDS.

\end{tcolorbox}

\subsection{Chunking}
\label{sec:appendix_llms_chunking}
We chunk long documents where applicable with overlap between chunks. For open source LLMs, we use chunk size of 2,048 tokens with a 128 token overlap. For commercial LLMs, we use chunk size of 8,192. We aggregate the labels for each word, and assign label 1 if there is a segmentation in any of the overlapping parts of the chunk.

\section{Licenses}
\label{app:license}

We list the licenses of the data and tools used in this work below:

\begin{itemize}
    \item \textbf{Creative Commons Attribution Share Alike 4.0}: BAREC Corpus~\cite{elmadani-etal-2025-large}.
    \item \textbf{MIT License}: CAMeL Tools~\citep{obeid-etal-2020-camel} and CAMeLBERT~\cite{inoue-etal-2021-interplay}.
    \item \textbf{Qwen License}: Qwen2.5-72B-Instruct~\cite{qwen2,qwen2.5}
    \item \textbf{Apache License 2.0}: Fanar-2-27B-Instruct~\cite{fanarteam2026fanar20arabicgenerative}, Jais-2-70B-Chat~\cite{anwar2026jais2familyarabiccentric}, Qwen3.6-27B~\cite{qwen3.6-27b}.
\end{itemize}

\clearpage
\onecolumn
\section{Genre-Level Results}
\label{app:genre-res}









\begin{table*}[th!]
\centering
\small
\setlength{\tabcolsep}{2.2pt}
\begin{tabular}{lcccc|cccc|cccc|cccc}
\toprule
 & \multicolumn{4}{c}{\textbf{NoPnx-NP}} & \multicolumn{4}{c}{\textbf{NoPnx-PA}} & \multicolumn{4}{c}{\textbf{NP}} & \multicolumn{4}{c}{\textbf{PA}} \\
\cmidrule(lr){2-5}
\cmidrule(lr){6-9}
\cmidrule(lr){10-13}
\cmidrule(lr){14-17}
& \textbf{SaT}
& \textbf{Gemini}
& \textbf{Parser}
& \textbf{CB}
& \textbf{SaT}
& \textbf{Gemini}
& \textbf{Parser}
& \textbf{CB}
& \textbf{SaT}
& \textbf{Gemini}
& \textbf{Parser}
& \textbf{CB}
& \textbf{SaT}
& \textbf{Gemini}
& \textbf{Parser}
& \textbf{CB} \\

\midrule
Children & 67.3 & 79.7 & 78.8 & \textbf{87.2} & 84.3 & 90.3 & 88.7 & \textbf{91.3} & 70.9 & 68.8 & 84.5 & \textbf{89.3} & 83.8 & 87.7 & 75.0 &\textbf{95.8} \\

Education & 56.6 & 61.6 & 75.2 & \textbf{89.9} & 67.6 & 66.3 & 85.3 & \textbf{92.7} & 55.5 & 58.5 & 93.3 & \textbf{96.1} & 64.5 & 64.8 & 83.1 &\textbf{96.8} \\

Literature & 53.0 & 75.0 & 73.7 & \textbf{80.7} & 65.7 & 78.4 & 82.6 & \textbf{85.8} & 68.4 & 64.1 & 85.8 &\textbf{88.5} & 71.2 & 72.7 & 67.9 & \textbf{91.0} \\

Media & 80.5 & 82.4 & 84.4 &\textbf{88.0} & 79.6 & 84.6 & 89.4 &\textbf{90.8} & 78.2 & 72.9 & 91.0 &\textbf{91.3} & 78.4 & 80.3 & 75.4 & \textbf{93.1} \\

Poetry & 3.5 & \textbf{100.0} & 61.5 & 72.4 & 90.1 & \textbf{100.0} & 99.2 & 98.2 & 3.5 & \textbf{99.0} & 62.2 & 70.7 & 90.1 & \textbf{100.0} & 99.2 & 99.0 \\

Politics & \textbf{86.3} & 68.2 & 66.8 & 84.8 & 87.3 & 82.9 & 89.1 & \textbf{92.2} & 93.6 & 49.4 & 86.7 & \textbf{93.7} & 91.4 &  93.3 & 85.4 & \textbf{95.5} \\

Religion & 31.9 & 82.4 & 76.9 & \textbf{84.6} & 39.5 & 83.0 & 75.2 & \textbf{83.7} & 53.0 & 80.8 & 96.0 & \textbf{96.8} & 50.7 & 80.8 & 53.7 & \textbf{96.8} \\

Wiki & 73.4 & 79.7 & 75.6 & \textbf{84.7} & 76.6 & 85.2 & 81.1 & \textbf{85.7} & 74.7& 75.6 & 90.2 & \textbf{92.3} & 75.7 & 77.2 & 54.3 & \textbf{91.9} \\\midrule\midrule

Overall & 53.0 & 75.3 & 76.5 & \textbf{85.9} & 63.7 & 79.6 & 82.8 & \textbf{88.4} & 61.8 &  69.5 & 90.7 & \textbf{93.2} & 66.8 & 76.6 & 82.8 & \textbf{95.2} \\
\bottomrule
\end{tabular}
\caption{Genre-level results in terms of F\textsubscript{1} on the Test set of {\AraSEG} across the four task variants. Best results are in bold. Parser denotes the Flat Orig Deprel Parser; CB denotes CAMeLBERT.}
\label{tab:genre-results}
\end{table*}

\section{Error Analysis}
\label{app:error-analysis}

\subsection{Genre-Level Punctuation- vs. Word-Based Errors}
\label{app:pnx-word-genre-error-analysis}

\begin{table*}[h!]
\centering
\small
\setlength{\tabcolsep}{3.2pt}
\begin{tabular}{l*{4}{cccc}}
\toprule
& \multicolumn{4}{c}{\textbf{NoPnx-NP}}
& \multicolumn{4}{c}{\textbf{NoPnx-PA}}
& \multicolumn{4}{c}{\textbf{NP}}
& \multicolumn{4}{c}{\textbf{PA}} \\
\cmidrule(lr){2-5}
\cmidrule(lr){6-9}
\cmidrule(lr){10-13}
\cmidrule(lr){14-17}
\textbf{Genre}
& \multicolumn{2}{c}{\textbf{Pnx}}
& \multicolumn{2}{c}{\textbf{Word}}
& \multicolumn{2}{c}{\textbf{Pnx}}
& \multicolumn{2}{c}{\textbf{Word}}
& \multicolumn{2}{c}{\textbf{Pnx}}
& \multicolumn{2}{c}{\textbf{Word}}
& \multicolumn{2}{c}{\textbf{Pnx}}
& \multicolumn{2}{c}{\textbf{Word}} \\
\midrule

Children
& 244 & (64\%) & 140 & (36\%)
& 203 & (78\%) & 58 & (22\%)
& 151 & (54\%) & 127 & (46\%)
& 117 & (85\%) & 21 & (15\%) \\

Education
& 551 & (54\%) & 463 & (46\%)
& 371 & (58\%) & 270 & (42\%)
& 207 & (62\%) & 129 & (38\%)
& 130 & (60\%) & 86 & (40\%) \\

Literature
& 610 & (63\%) & 360 & (37\%)
& 415 & (65\%) & 222 & (35\%)
& 327 & (55\%) & 270 & (45\%)
& 237 & (57\%) & 178 & (43\%) \\

Media
& 47 & (64\%) & 27 & (36\%)
& 36 & (62\%) & 22 & (38\%)
& 14 & (37\%) & 24 & (63\%)
& 17 & (46\%) & 20 & (54\%) \\

Poetry
& 1 & (2\%) & 65 & (98\%)
& 0 & (0\%) & 4 & (100\%)
& 1 & (1\%) & 70 & (99\%)
& 0 & (0\%) & 2 & (100\%) \\

Politics
& 83 & (77\%) & 25 & (23\%)
& 54 & (70\%) & 23 & (30\%)
& 76 & (78\%) & 21 & (22\%)
& 55 & (71\%) & 23 & (29\%) \\

Religion
& 504 & (81\%) & 121 & (19\%)
& 434 & (81\%) & 99 & (19\%)
& 28 & (56\%) & 22 & (44\%)
& 26 & (54\%) & 22 & (46\%) \\

Wiki
& 108 & (80\%) & 27 & (20\%)
& 98 & (78\%) & 27 & (22\%)
& 53 & (80\%) & 13 & (20\%)
& 59 & (87\%) & 9 & (13\%) \\

\midrule\midrule
Total
& 2,148 & (64\%)
& 1,228 & (36\%)
& 1,611 & (69\%)
& 725 & (31\%)
& 857 & (56\%)
& 676 & (44\%)
& 641 & (64\%)
& 361 & (36\%) \\

\bottomrule
\end{tabular}

\caption{CAMeLBERT punctuation- and word-based error distribution on the Test set of {\AraSEG} across genres.}

\label{tab:pnx-word-errors-all-settings}
\end{table*}

\clearpage
\subsection{Genre-Level Punctuation Errors}
\label{app:pnx-genre-error-analysis}

\begin{table*}[h!]
\centering
\small
\begin{tabular}{l*{13}{c}}
\toprule
\textbf{Genre}
& \textbf{\texttt{,}}
& \textbf{\texttt{.}}
& \textbf{\texttt{(}}
& \textbf{\texttt{)}}
& \textbf{\texttt{:}}
& \textbf{\texttt{"}}
& \textbf{\texttt{?}}
& \textbf{\texttt{-}}
& \textbf{\texttt{..}}
& \textbf{\texttt{;}}
& \textbf{\texttt{!}}
& \textbf{Other}
& \textbf{Total} \\
\midrule

\multicolumn{14}{c}{\textbf{NP}} \\
\midrule

Children
& 63\% & 1\% & 0\% & 1\%
& 6\% & 2\% & 0\% & 2\%
& 21\% & 1\% & 1\% & 3\%
& 18\% \\

Education
& 43\% & 21\% & 0\% & 2\%
& 19\% & 1\% & 1\% & 1\%
& 1\% & 5\% & 0\% & 7\%
& 24\% \\

Literature
& 72\% & 2\% & 0\% & 1\%
& 8\% & 0\% & 1\% & 1\%
& 0\% & 8\% & 1\% & 6\%
& 38\% \\

Media
& 93\% & 0\% & 0\% & 0\%
& 0\% & 0\% & 0\% & 0\%
& 0\% & 0\% & 0\% & 7\%
& 2\% \\

Poetry
& 0\% & 0\% & 0\% & 0\%
& 0\% & 0\% & 0\% & 0\%
& 0\% & 0\% & 0\% & 100\%
& 0\% \\

Politics
& 16\% & 75\% & 0\% & 8\%
& 1\% & 0\% & 0\% & 0\%
& 0\% & 0\% & 0\% & 0\%
& 9\% \\

Religion
& 79\% & 0\% & 0\% & 0\%
& 0\% & 7\% & 4\% & 0\%
& 0\% & 0\% & 0\% & 11\%
& 3\% \\

Wiki
& 85\% & 4\% & 0\% & 0\%
& 0\% & 2\% & 0\% & 0\%
& 0\% & 4\% & 0\% & 6\%
& 6\% \\

\midrule\midrule
Total
& 59\%
& 13\%
& 0\%
& 2\%
& 9\%
& 1\%
& 1\%
& 1\%
& 4\%
& 5\%
& 0\%
& 6\%
& 100\% \\

\midrule

\multicolumn{14}{c}{\textbf{PA}} \\
\midrule

Children
& 71\% & 3\% & 0\% & 1\%
& 1\% & 0\% & 0\% & 1\%
& 19\% & 1\% & 1\% & 3\%
& 18\% \\

Education
& 66\% & 2\% & 0\% & 1\%
& 19\% & 0\% & 0\% & 1\%
& 0\% & 6\% & 0\% & 6\%
& 20\% \\

Literature
& 76\% & 1\% & 0\% & 1\%
& 3\% & 0\% & 2\% & 1\%
& 0\% & 7\% & 0\% & 8\%
& 37\% \\

Media
& 88\% & 0\% & 0\% & 0\%
& 0\% & 0\% & 0\% & 0\%
& 6\% & 0\% & 0\% & 6\%
& 3\% \\

Poetry
& -- & -- & -- & --
& -- & -- & -- & --
& -- & -- & -- & --
& 0\% \\

Politics
& 11\% & 87\% & 0\% & 0\%
& 2\% & 0\% & 0\% & 0\%
& 0\% & 0\% & 0\% & 0\%
& 9\% \\

Religion
& 89\% & 0\% & 0\% & 0\%
& 0\% & 8\% & 0\% & 0\%
& 0\% & 0\% & 0\% & 4\%
& 4\% \\

Wiki
& 92\% & 3\% & 0\% & 0\%
& 0\% & 2\% & 0\% & 0\%
& 0\% & 3\% & 0\% & 0\%
& 9\% \\

\midrule\midrule
Total
& 70\%
& 9\%
& 0\%
& 1\%
& 5\%
& 1\%
& 1\%
& 1\%
& 4\%
& 4\%
& 0\%
& 5\%
& 100\%\\

\bottomrule
\end{tabular}

\caption{CAMeLBERT punctuation-level error distribution across genres on the Test set of {\AraSEG}.}
\label{tab:pnx-errors-by-genre}
\end{table*}

\subsection{Word-Based Errors}
\label{app:word-error-analysis}
\begin{table}[h!]
\small
\centering
\begin{tabular}{lcc|cc|cc|cc}
\toprule
\textbf{Error Category}
&  \multicolumn{2}{c}{\textbf{NoPnx-NP}} 
&  \multicolumn{2}{c}{\textbf{NoPnx-PA}} 
&  \multicolumn{2}{c}{\textbf{NP}}
&  \multicolumn{2}{c}{\textbf{PA}}\\
\midrule
Paragraph Boundary
& 26 & (16\%)
& -- & --
& 28 & (18\%)
& -- & -- \\

Lexical/Discourse Ambiguity
& 90 & (56\%)
& 109 & (68\%)
& 95 & (59\%)
& 122 & (76\%) \\

Verse Boundary
& 21 & (13\%)
& 23 & (14\%)
& 18 & (11\%)
& 8 & (5\%) \\

Dialogue Transition
& 16 & (10\%)
& 23 & (14\%)
& 18 & (11\%)
& 28 & (18\%) \\

Other
& 7 & (4\%)
& 5 & (3\%)
& 1 & (1\%)
& 2 & (1\%) \\
\midrule\midrule
Total
& 160 & 
& 160 & 
& 160 & 
& 160 &  \\
\bottomrule
\end{tabular}
\caption{Manual analysis of sampled CAMeLBERT word errors across the four task settings. Paragraph Boundary denotes errors aligned with paragraph breaks; Lexical/Discourse Ambiguity includes unclear boundaries involving lexical, syntactic, connective, or discourse cues; Verse Boundary covers verse-structured text; Dialogue Transition covers quoted or conversational speech.}
\label{tab:word-error-analysis}
\end{table}

\end{document}